\documentclass[11pt,draftclsnofoot,onecolumn]{IEEEtran}

\usepackage{graphicx}
\usepackage{epstopdf}
\usepackage{epsfig}
\usepackage{amsmath}
\usepackage{bbm}
\usepackage{bm}
\usepackage{amssymb}
\usepackage{wrapfig}
\usepackage{times,color}
\usepackage{cite,footnote,xspace,syntonly,algorithm,algorithmic,bm}
\usepackage[caption=false,font=footnotesize]{subfig}

\usepackage{amsfonts}

\usepackage{epstopdf}

\input{mysymbol.sty}

\def \cL {\mathcal{L}}
\def \cF {\mathcal{F}}

\def \ubL {\underline{\bf L}}

\def \ubW {\underline{\bf W}}
\def \ubX {\underline{\bf X}}
\def \ubY {\underline{\bf Y}}

\def \bY {{\bf Y}}

\def \bX {{\bf X}}
\def \bA {{\bf A}}
\def \bV {{\bf V}}
\def \bW {{\bf W}}
\def \bF {{\bf F}}

\def \bI {{\bf I}}

\def \bM {{\bf M}}
\def \bU {{\bf U}}
\def \bV {{\bf V}}

\def \bD {{\bf D}}

\def \bW {{\bf W}}
\def \bH {{\bf H}}
\def \bL {{\bf L}}

\def \bx {{\bf x}}

\def \be {{\bf e}}

\def \by {{\bf y}}
\def \ba {{\bf a}}

\def \bSigma {{\boldsymbol \Sigma}}

\def \bPhi {{\boldsymbol \Phi}}
\def \bphi {{\boldsymbol \phi}}
\def \balpha {{\boldsymbol \alpha}}

\def \bgamma {{\boldsymbol \gamma}}

\def \bbeta {{\boldsymbol \beta}}

\def \tr {\text{tr}}

\def \diag {\text{diag}}

\long\def\symbolfootnote[#1]#2{\begingroup
\def\thefootnote{\fnsymbol{footnote}}
\footnote[#1]{#2}\endgroup} \psfull

\begin{document}
\title{\huge Tracking Tensor Subspaces with Informative Random Sampling for Real-Time MR Imaging$^\dag$}

\author{{\it Morteza Mardani, Georgios~B.~Giannakis, and Kamil Ugurbil}$^\ast$}

\markboth{IEEE TRANSACTIONS ON MEDICAL IMAGING (SUBMITTED)}
\maketitle \maketitle  \symbolfootnote[0]{$\dag$ Work in this paper
was supported by the NIH grant 1R01GM104975-01, and NSF grants 1514056 and 1500713. Parts of the
paper was presented in the {\it 36th IEEE Engineering in Medicine and Biology Conference (EMBC)}, Chicago, IL, September 2014; and the {\it 23th ISMRM Annual Meeting and Exhibition}, Toronto, Ontario, May 2015.}

\symbolfootnote[0]{$\ast$ M. Mardani is with the Stanford University, School of Medicine, 875 Blake Wilbur Drive Stanford, CA 94305; Email: \texttt{morteza@stanford.edu.} G. B. Giannakis is with the Dept.
of ECE and the Digital Technology Center, University of
Minnesota, 200 Union Street SE, Minneapolis, MN 55455. Tel/fax:(612) 625-4287/625-2002; Email: \texttt{georgios@umn.edu.} K. Ugurbil is also with the Center for Magnetic Resonance Research and the University of Minnesota Medical School, 2021 Sixth street SE, Minneapolis, MN 55416, Tel:(612)626-9591; \texttt{kamil@cmrr.umn.edu}}

\vspace*{-90pt}
\begin{center}
\small{\bf Submitted: }\today
\end{center}


\thispagestyle{empty}\addtocounter{page}{-1}
\begin{abstract}
\vspace{-3mm}
Magnetic resonance imaging (MRI) nowadays serves as an important modality for diagnostic and therapeutic guidance in clinics. However, the {\it slow acquisition} process, the dynamic deformation of organs, as well as the need for {\it real-time} reconstruction, pose major challenges toward obtaining artifact-free images. To cope with these challenges, the present paper advocates a novel subspace learning framework that permeates benefits from parallel factor (PARAFAC) decomposition of tensors (multiway data) to low-rank modeling of temporal sequence of images. Treating images as multiway data arrays, the novel method preserves spatial structures and unravels the latent correlations across various dimensions by means of the tensor subspace. Leveraging the spatio-temporal correlation of images, Tykhonov regularization is adopted as a rank surrogate for a least-squares optimization program. Alteranating majorization minimization is adopted to develop online algorithms that recursively procure the reconstruction upon arrival of a new undersampled $k$-space frame. The developed algorithms are {\it provably convergent} and highly {\it parallelizable} with lightweight FFT tasks per iteration. To further accelerate the acquisition process, randomized subsampling policies are devised that leverage intermediate estimates of the tensor subspace, offered by the online scheme, to {\it randomly} acquire {\it informative} $k$-space samples. In a nutshell, the novel approach enables tracking motion dynamics under low acquisition rates `on the fly.' GPU-based tests with real {\it in vivo} MRI datasets of cardiac cine images corroborate the merits of the novel approach relative to state-of-the-art alternatives. 
\end{abstract}


\vspace*{-5pt}

\vspace{-3mm}
\begin{keywords}
\vspace{-3mm}
Real-time MRI, PARAFAC decomposition, tensor subspace learning, low-rank, randomized subsampling. 
\end{keywords}

\vspace{-0.15cm}

\section{Introduction}
\label{sec:intro}
Since its inception in the 70s, magnetic resonance imaging (MRI) has emerged as a premier tool for biomedical imaging~\cite{zhi2000principles}. In recent years, MRI has also shown tremendous potential for dynamic processing. Through different protocols, dynamic MRI is able to provide images of tissues, perfusion, diffusion, spectroscopy, and susceptibility, both qualitatively as well as quantitatively in 3D high resolution, and in real time for every patient. The abundance of diverse data across time offers unprecedented opportunities to understand, diagnose, and treat diseases. Nevertheless, with such big blessings come big challenges, including: (c1) acquisition that is rather slow, allowing only for a limited amount of data to be collected per time slot to ensure that {\it motion} is frozen; and (c2) {\it real-time} image reconstruction is limited by the associated tradeoffs between speed and accuracy. Consequently, only low-resolution dynamic images can be acquired by state-of-the-art real-time MRI scanners.

Ample research has been carried out over the last decade to accelerate the dynamic MRI scanning process~\cite{lustig2006kt,huang2005kt,otazo2012combination,kozerke2004accelerating,jung2009k,lingala2011accelerated,pedersen2009k}. One way or another, existing works exploit the spatio-temporal correlation of MR images. Compressive sampling (CS) has been widely employed to leverage the parsimonious nature of data in a proper transform domain by means of sparsity and low rank regularization. The noteworthy representatives include {\it k-t} SPARSE~\cite{lustig2006kt}, {\it k-t} GRAPPA~\cite{huang2005kt}, {\it k-t} SPARSE SENSE \cite{otazo2012combination}, {\it k-t} BLAST~\cite{kozerke2004accelerating}, {\it k-t} FOCUSS~\cite{jung2009k}, {\it k-t} SLR~\cite{lingala2011accelerated}, and {\it k-t} PCA~\cite{pedersen2009k}. They however rely on batch data processing, and typically non-smooth optimization modules which are relatively slow and demand high computational and storage resources for high resolution imaging. While batch processing is affordable for diagnostic purposes, image-guided therapeutic and surgical navigations demand real-time tracking for the orgrans of interest. There is a handful of studies on real-time MRI reconstruction that rely either on Kalman filtering, or, online compressive sampling; see e.g.,~\cite{sumbul2009practical,majumdar2012compressed}. Kalman filtering based techniques~\cite{sumbul2009practical} capture motion dynamics via state-space models, and end up with fast but low-quality reconstruction. CS-based methods such as \cite{majumdar2012compressed} build on the motion sparsity and yield higher quality images but they are comparatively slow. 



Aiming at fast and enhanced quality reconstruction, the present paper brings forth a novel tensor (multiway data array) subspace learning (TSL) framework that unravels the latent correlation structure of the MRI data stream. MR images comprise a multiway array with $x,y,z$ coordinates, as well as time and possibly the coil dimension for parallel imaging. For a general linear observation of an $M$-way array we postulate a low-rank model based on the parallel factor (PARAFAC) decomposition~\cite{kolda_tutorial} of tensors that summarizes the tensor latent subspace in $M-1$ factor matrices. A Tykhonov regularizer is adopted as a rank surrogate that regularizes a least-squares (LS) fitting cost to estimate the subspace and consequently interpolate the missing data. Broadening the scope of our precursor works in~\cite{mardani2014subspace} and \cite{MMG13-anomalography}, and leveraging the decomposable structure of the sought optimization formulation, stochastic alternating minimization is adopted to develop iterative solvers for which the acquisition time coincides with the iteration index. Upon acquisition, the new datum with partial $k$-space data is first projected onto the latest subspace estimate, and the mismatch refines the subspace. The resulting procedure boils down to lightweight iterates with parallelized computations that suit GPU implementation for high-resolution imaging. In addition, the resulting subspace sequence is provably convergent to the stationary point set of the batch objective.

For the possibly parallel MRI, reconstruction schemes are introduced that either interpolate the misses in the $k$-space, or, directly retrieve in the image domain pursuing a tomographic approach. The proposed schemes offer real-time reconstruction of MR images `on the fly.' Furthermore, the online subspace estimates, are utilized to devise a data-driven $k$-space subsampling rule to further accelerate the acquisition by collecting the most {\it informative} features for reconstruction. Specifically, inspired by randomized linear algebra approaches~\cite{mahoney2011randomized}, we put forth a novel importance score that ranks the $k$-space entries to be acquired in the next frame, according to their coherence level with the latest tensor subspace. GPU-based simulated tests with two different {\it in vivo} cardiac cine MR image datasets corroborate the effectiveness of the novel reconstruction schemes in terms of speed, and motion tracking relative to {\it k-t} FOCUSS~\cite{jung2009k} and differential CS~\cite{majumdar2012compressed}. Last but not least, the scope of the proposed framework goes beyond dynamic MRI, and can indeed cater to other `big data' inference tasks encountered with different medical imaging modalities.

The rest of this paper is organized as follows. Section~\ref{sec:prelim} introduces preliminaries on tensor PARAFAC as well as rank regularization, and advocates a model to arrive at a generic optimization setup for reconstruction. Section~\ref{sec:subs_tracking} then develops iterative solvers to track the tensor subspace. Subsequently, Section~\ref{sec:dynamic_mri} focuses on dynamic and parallel MRI settings, where two reconstruction schemes are proposed to either interpolate misses in the $k$-space or image domain in a tomographic manner. Adaptive random subsampling to further accelerate MR acquisition process is the subject of Section~\ref{sec:adapt_sampling}. Finally, real-data tests are reported in Section~\ref{sec:sim}, while conclusions are drawn in Section~\ref{sec:conc}.

\noindent{\it Notation}: Bold uppercase (lowercase) letters will denote matrices (column vectors with their entries in parenthesis), and calligraphic letters will be used for sets. Tensors or multi-way arrays are denoted by bold, underlined uppercase letters. Operators $(\cdot)^{\top}$, $(\cdot)^{*}$, $(\cdot)^{\mathsf{H}}$, $\rm{tr}(\cdot)$, $\mathbb{E}[\cdot]$, $\sigma_{\max}(\cdot)$, $\odot$, and $\circ$ will denote transposition, complex conjugate, Hermitian, matrix trace, statistical expectation, maximum singular value, Hadamard product, and outer product, respectively; $|\cdot|$ will be used for the cardinality of a set, and the magnitude of a scalar. The positive semidefinite matrix $\mathbf{M}$ will be denoted by $\bbM\succeq\mathbf{0}$. The $\ell_p$-norm of $\bx \in \mathbb{R}^n$ is $\|\bx\|_p:=(\sum_{i=1}^n |x_i|^p)^{1/p}$ for $p \geq 1$. For two matrices $\bM,\bU \in \mathbb{R}^{n \times p}$,
$\langle \bM, \bU \rangle := \rm{tr(\bM^{\top} \bU)}$ denotes their trace inner 
product, and $\|\bM\|_F:=\sqrt{\tr(\bM\bM^{\top})}$ is the Frobenius norm. The $n \times n$ identity matrix will be represented by $\bI_n$, while $\mathbf{0}_{n}$ will stand for the $n \times 1$ vector of all zeros, $\mathbf{0}_{n \times p}:=\mathbf{0}_{n} \mathbf{0}^{\top}_{p}$, and $[n]:=\{1,2,\ldots,n\}$. Also, $\be_i$ denotes the canonical vector with one at $i$-th entry and zero elsewhere, while the operators ${\rm vec}$ and ${\rm unvec}$ stack the columns of a matrix on top of each other, and vice versa.

\section{Preliminaries and Problem Statement}
\label{sec:prelim}
As modern and massive datasets become increasingly complex and heterogeneous, in many application setups one encounters data structures indexed by three or more variables giving rise to a tensor, instead of just two variables indexing data organized in a matrix. A few examples of time-indexed, medical tensor
data include~\cite{kolda_completion}: (i) images acquired in parallel MRI across various coils, as well as snapshots across time and patients, collected in a five-dimensional array with (phase encoding, frequency encoding, coil, time, patient); and (ii) Electroencephalograms (EEGs), where the signal of each electrode is a
time-frequency matrix; thus, data from multiple channels is three-dimensional (temporal, spectral, 
and spatial) and may be incomplete if electrodes become loose or disconnected for a period of time.

\subsection{Low-rank PARAFAC decomposition}
\label{subsec:parafac_decomp}
For multiple, say $M\geq 2$, vectors $\ba_m \in\mathbb{C}^{N_m\times 1}$, the outer product $\ba_1 \circ \ldots \circ \ba_M $ is an $N_1 \times \ldots \times N_M$ rank-one $M$-way array with $(n_1,\ldots,n_M)$-th entry given by 
$\Pi_{m=1}^M a_{n_m,m}$, where $a_{n_m,m}$ is the $n_m$-th entry of $\ba_m$. This generalizes the matrix case ($M=2$), where $\ba_1 \circ \ba_2 =\ba_1 \ba_2^{\top}$ is a rank-one matrix. The rank of a tensor 
$\underline{\bX}$ is defined as the minimum number of outer products required to 
synthesize $\underline{\bX}$. The PARAFAC model is arguably the most basic model because of its direct relationship to tensor rank. Specifically, it is natural to form an $R$-rank approximation of tensor $\underline{\bX} \in \mathbb{C}^{N_1 \times \ldots \times N_M}$ as
\begin{align}
\underline{\bX} \approx \sum_{r=1}^R \ba_r^{(1)} \circ \ldots \circ \ba_r^{(M)} . \label{eq:parafac}
\end{align}
When the approximation is exact, \eqref{eq:parafac} is the PARAFAC decomposition of $\underline{\bX}$. Accordingly, the minimum value $R$ for which the exact decomposition is possible is (by definition) the rank of $\underline{\bX}$. Different from the matrix case, there is no straightforward algorithm to determine the rank of a given tensor, a problem that is known to be NP-hard~\cite{kolda_tutorial}.

\begin{figure}[t]
\centering
\includegraphics[scale=0.4]{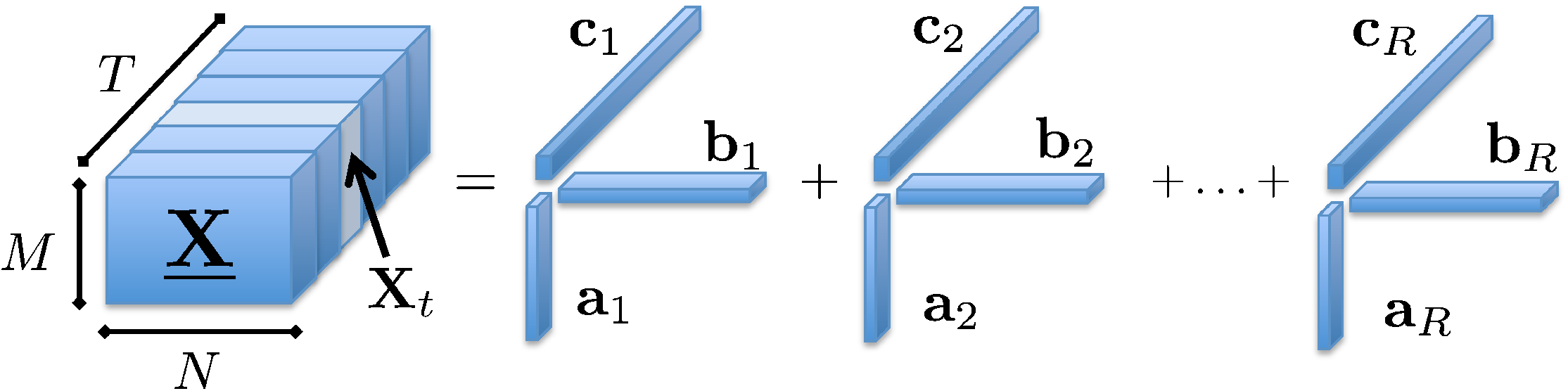}
\caption{A rank-$R$ PARAFAC decomposition of the three-way tensor $\underline{\mathbf{X}}$.}
\label{fig:Fig_1}
\end{figure}

With reference to \eqref{eq:parafac}, introduce the factor matrices~$\bA_m:=[\ba_1^{(m)},\ldots, \ba_R^{(m)} ] \in \mathbb{C}^{N_m \times R},~i\in [M]$, and let $\ubX_{\ell}^{(k)},\ \ell \in [N_k]$ denote the $\ell$-th slice of $\underline{\bX}$ along its $k$-th mode, such that $\ubX_{\ell}^{(k)}(n_1,\ldots,n_{k-1},n_{k+1},\ldots,n_M)$ $=\underline{\bX}(n_1,\ldots,\ell,\ldots,n_M)$. This $\ell$-th slice can then be expressed as (cf.~\eqref{eq:parafac})
\begin{align}
\ubX_{\ell}^{(k)} \hspace{-1mm}=\hspace{-1mm} \sum_{r=1}^R \bgamma_{\ell,r}^{(k)} \ba_r^{(1)} \circ \ldots \circ \ba_r^{(k-1)} \circ \ba_r^{(k+1)}  \circ \ldots  \circ \ba_r^{(M)},\quad \ell=1,\ldots,N_k \label{eq:slice_t}
\end{align}
where $\bgamma_{\ell} \in \mathbb{C}^R$ denotes the $\ell$-th row of $\bA_k$ whose $r$-th column is $\ba_r^{(k)}$. To gain intuition, imagine a three-way data array ($M=3$), where the slices form matrices, and thus the $\ell$-th slice across the tube dimension is given by $\bX_{\ell}=\sum_{r=1}^R \gamma_{\ell,r}^{(3)} \ba_r^{(1)} {\ba_r^{(2)}}^{\top}$. It is apparent that a slice $\bX_{\ell}$ can be represented as a linear combination of $R$ rank-one matrices $\{\ba_r^{(1)} {\ba_r^{(2)}}^{\top}\}_{r=1}^R$, which constitute the bases for the tensor tube subspace. In the same manner, one can argue that for a general $M$-th order tensor, the rank-one tensors $\{\ba_r^{(1)} \circ \ldots \circ \ba_r^{(k-1)} \circ \ba_r^{(k+1)}  \circ \ldots  \circ \ba_r^{(M)}\}_{r=1}^R$ form the bases for the tensor subspace along the $k$-th mode. Likewise, $R$-dimensional vector $\bgamma_{\ell}^{(k)}$ collects the tensor principal components.

This study aims at discovering the latent subspace that is captured by the matrices $\{\bA_m\}_{m=1,m\neq k}^M$. This will become handy later. Given $\underline{\mathbf{X}}$, under mild conditions, matrices $\{\bA_m\}_{m=1}^M$ are unique up to a common column permutation and scaling (meaning that PARAFAC is essentially identifiable for $M\geq 3$); see e.g.~\cite{tBS02,jk77laa}. It is worth commenting that the factor matrices $\{\bA_m\}_{m=1}^M$ are not necessarily orthogonal, and may even be rank deficient. Thanks to its essential uniqueness, PARAFAC has become the model of choice when one is 
primarily interested in revealing latent structure in multiway data arrays. Considering the analysis of a
dynamic social network for instance, each of the rank-one factors
could correspond to communities that e.g., persist or form and dissolve dynamically across time~\cite{brian_tsp_social_nets_april2016}.

PARAFAC's link with the tensor rank can be used to postulate low-rank tensor models. However, as mentioned earlier even finding the tensor rank is an NP-hard problem. Parallel to the matrix nuclear-norm, tractable surrogates can be adopted for the tensor PARAFAC-rank that approximates the rank through the norm of its factor matrices. One such surrogate for $M=3$ is introduced in our companion work~\cite{juan_tensor_tsp_2013}. Generalizing~\cite{juan_tensor_tsp_2013} to multi-way arrays, we adopt the following rank surrogate for the tensor rank of $\ubX$
\begin{align}
\mathcal{Q}(\ubX):=\min_{\{\bA_i \in \mathbb{C}^{N_i \times R}\}_{i=1}^M}~~\frac{1}{M} \sum_{i=1}^M \|\bA_i\|_F^2 \quad {\rm s. to} \quad \underline{\bX} = \sum_{r=1}^R \ba_r^{(1)} \circ \ldots \circ \ba_r^{(M)}. \label{eq:rank_reg}
\end{align}
The proof that $\mathcal{Q}(\ubX)$ induces low tensor rank follows the main ideas in~\cite{juan_tensor_tsp_2013} for $M=3$, but it is omitted here due to space limitation. Nonetheless, the formal claim is given next.

\begin{lemma}\label{lem:lemma_1}
If $\sigma_r:=\Pi_{m=1}^{M} \|\ba_r^{(m)}\|$ denotes the $r$-th singular value of the $M$-way tensor $\ubX$, it then holds that
\begin{align}
\mathcal{Q}(\ubX)=\bigg( \sum_{r=1}^R |\sigma_r|^{2/M}    \bigg)^{2/M}. \nonumber
\end{align}
\end{lemma}
\noindent Lemma \ref{lem:lemma_1} asserts that $\mathcal{Q}(\ubX)$ promotes sparsity across the singular values of $\ubX$. Note that for $M=2$ the adopted regularizer boils down to the well-known matrix nuclear norm~\cite{fazel_phdthesis}.


\subsection{Subsampled data model}
\label{subsec:model}
In various application domains, the physical data of interest collected in a tensor vary slowly and can thus be approximated by a stationary process. Accordingly, consider that the $M$-way tensor process $\{\ubL_t\}$ lives in a low-dimensional subspace $\cL$. From this process, undersampled observations $\{\by_t \in \mathbb{C}^{L_t} \}$ that are streaming over time obey the model
\begin{align}
y_t^{(\ell)} = \langle \ubL_t, \ubW_{t}^{(\ell)} \rangle + v_t^{(\ell)},~~~\ell=1,\ldots,L_t \label{eq:obs_model}
\end{align}
where the projection tensor $\ubW_t^{(\ell)} \in \mathbb{C}^{N_1 \times \ldots \times N_{M-1}}$ sketches (meaning subsamples) $\ubL_t$, and $v_{t}^{(\ell)}$ accounts for the errors and unmodeled dynamics. This model hits several modern application domains such as dynamic MRI, where the ground-truth sequence of images form a three-way data cube, and per time slot $t$ a small subset of $k$-space data in the Fourier domain are acquired; further details are provided in Section~\ref{sec:dynamic_mri}.


Assume temporarily that one has only access to a batch of observations \eqref{eq:obs_model} during the time horizon $t \in [1,T]$. Collect the $(M-1)$-way tensors $\{\ubL_t\}_{t=1}^T$ into a larger $M$-way tensor $\ubL$ with the $M$-th mode representing time. Low dimensionality implies $\ubL$ has low PARAFAC rank. All in all, we wish to identify $\{\ubL_t\}_{t=1}^T$ given the observations $\{\by_t\}_{t=1}^T$ along with the projection tensors $\{\ubW_{t}^{(\ell)}\}_{t=1}^T$, assuming $\ubL$ is a low-rank tensor. Letting $\bgamma_t \in \mathbb{C}^R$ denote the $t$-th row of $\bA_M[t]$, a natural estimator $\hat{\ubL}$ is 
\begin{align}
{\rm (P1)} ~~~~~~~\hat{\ubL}=&\arg\min_{\{\bA_m\}_{m=1}^M,\{\bgamma_t\}_{t=1}^T}~\frac{1}{2} \sum_{t=1}^T\sum_{\ell=1}^{L_t} \big(y_{t}^{(\ell)} - \langle \ubL_t,\ubW_{t}^{(\ell)}  \rangle \big)^2 + \frac{\lambda}{2} \sum_{m=1}^M \|\bA_m\|_F^2  \nonumber \\
&\quad {\rm s.~to} \quad ~~~ \ubL_t = \sum_{r=1}^{R} \gamma_{t,r} \ba_r^{(1)} \circ \ldots \circ \ba_r^{(M-1)},\quad t=1,\ldots,T \nonumber
\end{align}
which fits the data to the postulated model \eqref{eq:obs_model} in the LS sense, and promotes low tensor rank through the regularizer $\sum_{m=1}^M \|\bA_m\|_F^2$. Tunning $\lambda$, controls the desired rank level. Note all the data and optimization variables in (P1) are complex valued.

In big data settings, the ambient dimensions $\{N_m\}_{m=1}^{M-1}$ can be quite large, and the tensor slices can be streaming over time with possibly $N_M \rightarrow \infty$ streams collected over time. Before delving into online solvers of (P1) for streaming observations, a couple of noteworthy properties of (P1) are in order. First, the rank regularization avoids the scaling ambiguity associated with the multilinear terms as stated next.


\begin{lemma}\label{lem:lemma_2}
Every stationary point of (P1) returns a tensor subspace with equal norm bases; that is,~$\|\ba_r^{(m)}\|=\|\ba_{r}^{(m')}\|,\forall m,m' \in [M-1]$, and $r \in [R]$. 
\end{lemma}

\begin{IEEEproof}~It readily follows by equating the gradient of (P1)'s objective w.r.t. $\ba_r^{(m)}$~(see also~\eqref{eq:gradient_a}) to zero, and taking the inner product of both sides with $\ba_r^{(m)}$. 
\end{IEEEproof}

Equal norm bases fix the scaling ambiguity inherent to the PARAFAC model. It also implies that the tensor singular values in Lemma~\ref{lem:lemma_1} are simply expressed as $\sigma_r=\|\ba_r^{(1)}\|^M$.

For large-scale inference tasks with $\prod_{m=1}^{M-1} N_m$ large solving (P1) incurs prohibitive complexity and storage to implement in batch mode. In addition, certain applications demand real-time processing upon acquisition of a new datum based on the past and current data, namely $\{y_{\tau}^{(\ell)},\ell \in [L_t]\}_{\tau=1}^t$. In essence, (P1) involves $R(N_1+\ldots+N_{M-1}+t)$ variables associated with the low-rank components, which can grow prohibitively with $t$, and eventually exceed the storage and computational limits. These obstacles press the need for online iterative solvers that can acquire data sequentially and perform simple update tasks. The ensuing section introduces machinery to arrive at such efficient online solvers.

\section{Tracking Tensor Subspace}
\label{sec:subs_tracking}
As elaborated in Section~\ref{subsec:model} the low-rank tensor $\ubL_t$ lies in a low-dimensional subspace $ \cL \subset \mathbb{C}^{N_1 \times \ldots \times N_{M-1}}$. With reference to PARAFAC-rank, $\cL$ is characterized by a small number $R$ of rank-one $(M-1)$-way arrays $\{ \ba_r^{(1)} \circ \ldots \circ \ba_r^{(M-1)}\}_{r=1}^R$, captured by the factor matrices~$\{\bA_m\}_{m=1}^{M-1}$. Learning these time-invariant factor matrices is the first step towards reconstructing the low-rank tensor of interest. With the streaming observations, at $t$-th acquisition time one is given $T=t$ data snapshots, and accordingly (with a slight abuse of notation) one is motivated to recast (P1) in the separable form
\begin{align}
{\rm (P2)}~~~~~~~~~\min_{\{\bA_m\}_{m=1}^{M-1}}~&\frac{1}{2t} \sum_{\tau=1}^t \min_{\bgamma_{\tau}} \Bigg\{ \sum_{\ell=1}^{L_t} \Big(y_{\tau}^{(\ell)} -  \sum_{r=1}^{R} \gamma_{\tau,r} \langle \ba_r^{(1)} \circ \ldots \circ \ba_r^{(M-1)},\ubW_t^{(\ell)}  \rangle  \rangle \Big)^2  \nonumber\\ & \hspace{4.5cm}  + \frac{\lambda}{2} \|\bgamma_{\tau}\|^2  \Bigg\} + \frac{\lambda}{2t} \sum_{m=1}^{M-1} \|\bA_m\|_F^2. \nonumber
\end{align}
Apparently, finding the optimal solution of the nonconvex program (P2) becomes computationally challenging especially for $M$ large. Hence, approximations that can afford simple iterative updates while approaching the optimal solution are well motivated. One such approximation for online rank minimization leveraging the separable nuclear-norm regularization~\eqref{eq:rank_reg} was introduced in~\cite{MMG13-anomalography} for matrices~($M=2$), in the context of unveiling network anomalies, and in~\cite{mardani2014subspace} for imputation of three-way tensors. Building on~\cite{mardani2014subspace} and \cite{MMG13-anomalography}, online solvers are developed next for the general $M \geq 3$ case.

\subsection{Stochastic alternating minimization}
\label{subsec:alt_min}
Towards deriving a real-time, computationally efficient, and
recursive solver of (P2), an alternating-minimization (AM)
method is adopted in which iterations coincide with the index $t$ of data acquisition. In accordance with (P2), consider the instantaneous regularized LS cost
\begin{align}
&f_{t}(\{\bA_i\}_{i=1}^{M-1};\bgamma_{t}) := \frac{1}{2}  \sum_{\ell=1}^{L_{t}} \Big(y_t^{(\ell)} -  \sum_{r=1}^{R} \gamma_{t,r} \langle \ba_r^{(1)} \circ \ldots \circ \ba_r^{(M-1)}, \ubW_{t}^{(\ell)}  \rangle \Big)^2 \nonumber \\ 
& \hspace{8cm} + (\lambda/2t) \sum_{m=1}^{M-1} \|\bA_m\|_F^2. \label{eq:f_t}
\end{align}
The iterative procedure adopted here consists of two major steps. The first step (S1)~relies on the recently updated subspace, namely $\{\bA_m[t-1]\}_{m=1}^{M-1}$, to solve the inner optimization, which yields the principal components $\bgamma_t=\arg\min_{\bgamma}~~f(\{\bA_i\}_{i=1}^{M-1};\bgamma)$. In the second step (S2),~the tensor subspace $\cL_t$ is updated by moving $\{\bA_m\}_{m=1}^{M-1}$ along the opposite direction of the gradient, namely~$-\nabla f_t(\{\bA_i\}_{i=1}^{M-1};\bgamma_t)$.

For (S1), collect $y_{t}^{(\ell)}$ in $\by_t \in \mathbb{C}^{L_t}$, and define matrix~$\bPhi_t := [\bphi_1^{(t)}, \ldots, {\bphi_{L_t}^{(t)}}]^{\top} \in \mathbb{C}^{L_t \times R}$, where $[\bphi_{t}^{(\ell)}]_r :=\langle \ba_r^{(1)} \circ \ldots \circ \ba_r^{(M-1)}, \ubW_{t}^{(\ell)} \rangle$. The projection of $\by_t$ onto the low-dimensional subspace $\cL_t$ is then obtained by solving the LS ridge-regression problem
\begin{align}
\bgamma_t = \arg\min_{\bgamma \in \mathbb{C}^R} \frac{1}{2} \|\by_t -    \bPhi_t \bgamma\|^2 + \frac{\lambda}{2} \|\bgamma\|^2 \nonumber
\end{align}
which admits the closed-form solution $\bgamma_{t}=(\bPhi_t^{\top}\bPhi_t+\lambda\bI_R)^{-1} \bPhi_t^{\top}\by_t$ that depends linearly on the subsampled data. To avoid the $R \times R$ matrix inversion, consider the SVD $\bPhi_t=\bU_t \bSigma_t \bV_t$ to end up with $\bgamma_t=\bV_t \bSigma_t^{-1}\bD_t \bU_t^{\top} \by_t$, where $\bD_t \in \mathbb{C}^{R \times R}$ is a diagonal matrix with $[\bD_t]_{i,i}=\sigma_i^2/(\sigma_i^2+\lambda_{*})$. 

%
%

The second step (S2) deals with updating the factor matrices given $\{\bgamma_{\tau}\}_{\tau=1}^t$ by solving 
\begin{align}
\{\bA_m[t]\}_{m=1}^{M-1} = \arg\min_{\{\bA_m\}_{m=1}^{M-1}} ~C_t(\{\bA_m\}_{m=1}^{M-1}):=\frac{1}{t} \sum_{\tau=1}^t f_{\tau}(\{\bA_m\}_{m=1}^{M-1};\bgamma_{\tau}). \label{eq:subspace_opt}
\end{align}
Apparently, \eqref{eq:subspace_opt} is a nonconvex program for $M \geq 3$ due to the multilinear terms in the LS cost, and is thus tough to solve optimally. To mitigate this computational challenge, consider the following quadratic approximant of $f_t$
\begin{align}
&\tilde{f}_t(\{\bA_m\}_{m=1}^{M-1};\bgamma_t) = f_t(\{\bA_m[t-1]\}_{m=1}^{M-1};\bgamma_t) \nonumber\\ &\hspace{1cm}+ \sum_{m=1}^{M-1}\langle \nabla_{\bA_m} f_t(\{\bA_m[t-1]\}_{m=1}^{M-1};\bgamma_t) ,\bA_m-\bA_m[t-1]\rangle + \frac{\alpha_t}{2} \sum_{m=1}^{M-1} \|\bA_m-\bA_m[t-1]\|_F^2   \nonumber
\end{align}
where $\alpha_t \geq \max_m \Big\{\sigma_{\max}\big[\nabla_{\bA_m}^2 f_t(\{\bA_m[t-1]\}_{m=1}^{M-1};\bgamma_t)\big] \Big\}$. With regards to the surrogate $\tilde{f}_t$, it is useful to recognize that it is locally tight, meaning that (i) $f_t(\{\bA_m[t-1]\}_{m=1}^{M-1};\bgamma_t) = \tilde{f}_t(\{\bA_m[t-1]\}_{m=1}^{M-1};\bgamma_t)$, and similarly $\nabla f_t(\{\bA_m[t-1]\}_{m=1}^{M-1};\bgamma_t) = \nabla \tilde{f}_t(\{\bA_m[t-1]\}_{m=1}^{M-1};\bgamma_t)$; and (ii) it upper bounds $f_t$, that is $f_t(\{\bA_m\}_{m=1}^{M-1};\bgamma_t) \leq \tilde{f}_t(\{\bA_m\}_{m=1}^{M-1};\bgamma_t)$,~for all $\bA_m \in \mathbb{C}^{N_m \times R}$,~and $m \in [M-1]$.

Apart from tightness, separability across factors is another attractive feature of $\tilde{f}_t$ because it allows for parallel implementation. Plugging in $\tilde{f}_t$ into the cost $C_t$ yields $\tilde{C}_t:=(1/t)\sum_{\tau=1}^t \tilde{f}_{\tau}$, the minimizer of which is obtained (after equating the gradient to zero) as
\begin{align}
\bA_m[t] =  \frac{1}{ \bar{\alpha}_t} \sum_{\tau=1}^{t} \alpha_{\tau} \Bigg\{ \bA_m[\tau-1] - \alpha_{\tau} \nabla_{\bA_m} f_{\tau}(\{\bA_m[\tau-1]\}_{m=1}^{M-1};\bgamma_{\tau})   \Bigg\} \nonumber
\end{align}
where $\bar{\alpha}_t:=\sum_{\tau=1}^t \alpha_{\tau}$. After rearranging terms one arrives at the recursion
\begin{align}
\bA_m[t] &= \big(\frac{1}{\bar{\alpha}_t}\big)  \underbrace{\sum_{\tau=1}^{t-1} \alpha_{\tau} \Bigg\{ \bA_m[\tau-1] - \alpha_{\tau}^{-1} \nabla_{\bA_m} f_{\tau}(\{\bA_m[\tau-1]\}_{m=1}^{M-1};\bgamma_{\tau})   \Bigg\} }_{:=\bar{\alpha}_{t-1}\bA_m[t-1]} \nonumber\\ &+ \big( \frac{\alpha_t}{\bar{\alpha}_t} \big)  \Bigg\{\bA_m[t-1] - \alpha_t \nabla_{\bA_m} f_t(\{\bA_m[t-1]\}_{m=1}^{M-1};\bgamma_{t}) \  \Bigg\} \nonumber \\
& = \bA_m[t-1] - (\bar{\alpha}_t)^{-1} \nabla_{\bA_m} f_t(\{\bA_m[t-1]\}_{m=1}^{M-1};\bgamma_{t}),\quad m \in [M-1]. \label{eq:recursion_gradient}   
\end{align}
Interestingly, \eqref{eq:recursion_gradient} is nothing but a single stochastic gradient descent step.

The gradient is separable across columns of $\bA_m$. Considering it w.r.t. each basis vector $\ba_r^{(m)}$ leads to the closed-form expression 
\begin{align}
&\nabla_{\ba_r^{(m)}} f_t(\bA_1,\ldots,\bA_{M-1}) = (\lambda/t) \ba_r^{(m)} \nonumber \\
&\hspace{2.3cm} -\sum_{\ell=1}^L \gamma_{t,r}^{*} \Big(y_{t}^{(\ell)} - \sum_{r=1}^R \gamma_{t,r}  \langle \ba_r^{(1)} \circ \ldots \circ \ba_r^{(M-1)} ,\ubW_{t}^{(\ell)}  \rangle \Big) \Big( \ubW_{t}^{(\ell)}  \times_{i=1,i \neq m}^{M-1} \ba_r^{(i)} \Big)^{*} \label{eq:gradient_a}
\end{align}
where $\times_i$ denotes the Tucker mode-$i$ product~\cite{kolda_tutorial}. It is also useful to recognize that the Hessian of $f_t$ admits the simple form
\begin{align}
&\nabla_{\ba_r^{(m)}}^2 f_t(\bA_1,\ldots,\bA_{M-1}) = (\lambda/t) \bI_{N_1} \nonumber \\
&\hspace{4cm} + |\gamma_{t,r}|^2 \sum_{\ell=1}^{L_t}  \Big(\ubW_{t}^{(\ell)}  \times_{i=1,i \neq m}^{M-1} \ba_r^{(i)} \Big) \Big( \ubW_{t}^{(\ell)}  \times_{i=1,i \neq m}^{M-1} \ba_r^{(i)} \Big)^{\mathsf{H}}.  \label{eq:hessian_a}
\end{align}
%


All in all, the gradient iterations for learning the tensor subspace proceed in parallel as follows
\begin{align}
\ba_r^{(m)}[t] = \ba_r^{(m)}[t-1] - \mu_t \nabla_{\ba_r^{(m)}} f_t(\{\bA_m[t-1]\}_{m=1}^{M-1};\bgamma_t), \quad r \in [R],~ m \in [M-1] \label{eq:gradeint_iteration}
\end{align}
where $\mu_t=\bar{\alpha_t}^{-1}$ denotes the step size. The resulting algorithm is listed under Table~\ref{tab:alg_lowrank_sparse_decomp}.

\subsection{Convergence analysis}
\label{subsec:cnvg}
Convergence of the first-order subspace iterates in Algorithm~\ref{tab:alg_lowrank_sparse_decomp} is granted following the analysis developed in~\cite{mairalonlinelearning} for convergence of dictionary learning, and our precursors in \cite{MMG13-anomalography,mardani2014subspace} establishing subspace convergence for imputation of two- and three-way arrays. The proof relies on martingale sequences, and in order to render the analysis tractable, it adopts the following assumptions:

{\it (as1)~The data stream $\{\by_t\}$ forms an i.i.d. random process that is uniformly bounded; and

(as2)~The subspace updates $\{\mathcal{L}[t]\}$ where $\mathcal{L}[t]:=\{\bA_m[t]\}_{m=1}^{M-1}$ lies in a compact set.}

With these assumptions, the convergence claim is formalized as follows. 

\begin{proposition}
If the subspace iterates $\{\cL[t]\}$ lie in a compact set, and the step-size sequence $\{\mu_t\}$ satisfies $(\mu_t)^{-1}:=\sum_{\tau=1}^t \alpha_{\tau} \geq ct,~\forall t$ for some $c>0$, where $c' \geq \alpha_t \geq \max_m \Big\{\sigma_{\max}\big[\nabla_{\bA_m}^2 f_t(\{\bA_m[t-1]\}_{m=1}^{M-1};\bgamma_t)\big] \Big\},~\forall t$ for some $c'>0$, then $\lim_{t \rightarrow \infty} \nabla C_t(\cL[t]) \rightarrow \mathbf{0}$, a.s.; i.e., the tensor subspace iterates $\{\cL[t]\}$ asymptotically coincide with the statioanry point set of the batch program (P2).  
\end{proposition}  
\vspace{2mm}

Note the step-size controlling constants $c$ and $c'$ determine the speed of convergence, and are chosen according to the observation parameters including the projection tensors $\{\ubW_t^{(\ell)}\}$. In particular, $c$ is related to the degree of curvature of the instantaneous loss $f_t$, and in a similar manner $c'$ is tied to the smoothness of $f_t$, and admits small values when the acquired observations end up with a smooth loss function having a small Lipschitz constant.

%
%

\begin{algorithm}[t]
\caption{Online rank-regularized tensor subspace learning} \small{
\begin{algorithmic}
	\STATE \textbf{input} 
	$\{y_{t}^{(\ell)},\ubW_t^{(\ell)},~\ell \in [L_t]\}_{t=1}^{\infty},\{\mu_t\}_{t=1}^{\infty},\lambda,R,M$.
    \STATE \textbf{initialize} $\{\bA_m[1]\}_{m=1}^M$ at random.
    \FOR {$t=1,2$,$\ldots$}
                \STATE \textbf{(S1) Projection coefficients update}                
                
                \STATE $[\bPhi_t]_{\ell,r}=\langle \ba_r^{(1)}[t-1] \circ \ldots \circ \ba_r^{(M-1)}[t-1], \ubW_{t}^{(\ell)} \rangle$,~~$\bPhi_t=\bU_t \bSigma_t \bV_t^{\top}$
                \STATE $\bD_t={\rm diag}\big[\sigma_1(\sigma_1^2+\lambda)^{-1},\ldots,\sigma_R(\sigma_R^2+\lambda)^{-1} \big]$ 
                
                

                \STATE $\bgamma_t=\bV_t \bSigma_t^{-1}\bD_t \bU_t^{\top} \by_t$
                
                \STATE $e_{t}^{(\ell)}:= y_{t}^{(\ell)}-\langle \bphi_{t}^{(\ell)}, \bgamma_t \rangle$

                \STATE \textbf{(S2) Parallel subspace update} $\big[(m,r) \in [M] \times [R] \big]$

                \STATE $\ba_r^{(m)}[t] = (1-\mu_t\lambda/t)\ba_r^{(m)}[t-1]  
 +\mu_t \gamma_{t,r}^{*} \Big(\sum_{\ell=1}^{L_t}  e_{t}^{(\ell)}  \big(\ubW_{t}^{(\ell)} \big)^{*} \Big) \times_{i=1,i \neq m}^{M-1} \big( \ba_r^{(i)}[t-1] \big)^{*} $
                 
                \RETURN  $\Big(\{\bA_m[t]\}_{m=1}^{M-1},\bgamma_t \Big)$
    \ENDFOR
\end{algorithmic}}
\label{tab:alg_lowrank_sparse_decomp}
\end{algorithm}

\noindent\textbf{Remark~1~[Parallelizable updates]:}~Implementing Algorithm~\ref{tab:alg_lowrank_sparse_decomp} per time instant $t$ involves updating the projection coefficients as well as the tensor subspace. The former mainly entails SVD computation of a $L_t \times R$ size matrix $\bPhi_t$, which incurs $\mathcal{O}(L_t R^2)$ operations. The latter is also nicely parallelizable across both the basis index $r$ and the factor index $m$, and hence $MR$ updates can be carried out simultaneously via parallel processors such as GPUs.

%
%
%

The ensuing section deals with application of the proposed tensor subspace learning scheme for reconstructing dynamic images in time-resolved MRI.

\section{Accelerating Dynamic MRI}
\label{sec:dynamic_mri}

Dynamic MRI acquires a low-spatial yet high-temporal resolution sequence of images, which renders a possibly sizable portion of measurements per snapshot inaccurate or missing. Fortunately, but the temporal correlation of images can be leveraged to interpolate these misses.

MRI typically uses a phased array of coils, each imaging a limited spatial region of the object. To begin, consider that single-coil MRI is used to acquire the ground-truth image sequence $\{\bL_t\}_{t=1}^T$ with the complex-valued image $\bL_t \in \mathbb{C}^{N_1 \times N_2}$ corresponding to $t$-th frame. Entries of $\bL_t$ record the collective magnetic field (including both phase and magnitude) induced per tissue voxel. Data corresponding to the $t$-th frame acquired in the frequency domain (hereafter referred to as $k$-space data) are then modeled as
\begin{align}
y_{t}^{(i,j)} = [\cF(\bL_t)]_{i,j} + v_{i,j},\quad \quad (i,j) \in \Omega_t \label{eq:mri_model}
\end{align}
where $\cF(\cdot)$ denotes the two-dimensional discrete Fourier transform (DFT) operator, and the set $\Omega_t \subset [N_1] \times [N_2]$ indexes the $k$-space data support. The acquisition time is clearly proportional to the sample count $\sum_{\tau=1}^t |\Omega_{\tau}|$, and it is desired to be as small as possible.

In what follows, a tensor imputation based approach is introduced first based on $k$-space correlations to interpolate the misses. For the multicoil scenario then a tomographic reconstruction scheme is devised that exploits the low rank directly on the image domain. To set up notation, the two-dimensional DFT operator $\cF(\cdot)$ can be written in compact matrix form as $\cF(\bX_t)=\bF_{l}\bX_t \bF_r$, with the left and right matrices $\bF_{l} \in \mathbb{R}^{N_1 \times N_1}$ and $\bF_r \in \mathbb{R}^{N_2 \times N_2}$, respectively, denoting orthonormal symmetric DFT matrices. The subsampling matrix in \eqref{eq:obs_model} is then $\bW_{t}^{(\ell)}=\bF_{\ell} \be_{i} \be_{j}^{\top} \bF_r$, and hence the projection corresponding to the $(i,j)$-th DFT coefficient is $\langle \bX_t,\bW_{t}^{(\ell)}  \rangle=[\cF(\bX_t)]_{i,j}$.

\subsection{$k$-space Interpolation}
\label{subsec:interpol_mri}
Recall that $\{\bL_t\}$ is the underlying image sequence of interest that lies in a low-rank tensor subspace $\cL$. Being {\it an orthonormal transform}, the Fourier operator preserves dimensionality, meaning that $\{\bX_t=\cF(\bL_t)\}$ lies in a linear tensor subspace, say $\cL_{F} \subset \mathbb{C}^{N_1 \times N_2}$ with ${\rm dim}(\cL_F) = {\rm dim}(\cL)$. Note also that the magnitude of eigenvalues for both the image and $k$-space data remains the same.

Accordingly, one can build on the low PARAFAC rank of the three-way tensor $\ubX$ to interpolate the misses from the present $k$-space entries. In this direction, given the partial $k$-space measurements
\begin{align}
y_t^{(i,j)} = x_t^{(i,j)} + v_t^{(i,j)}, \quad (i,j) \in \Omega_t \nonumber
\end{align}
we postulate a trilinear model $x_t^{(i,j)}=\langle \balpha_{i}, \boldsymbol{\beta}_{j}, \bgamma_t  \rangle$ with $\balpha_{i},\boldsymbol{\beta}_{j},\bgamma_{t}$ denoting rows of $\bA_1,\bA_2,\bA_3$, respectively. Choosing the sketching operator $\bW_t^{(i,j)}=\be_{i}\be_{j}^{\top}$, one can solve (P2) to learn the tensor factor matrices $\bA_1[t]$ and $\bA_2[t]$ `on the fly,' and subsequently interpolate to obtain $\hat{\bX}_t:=\bA_1[t] \diag(\bgamma_t) \bA_2^{\top}[t]$. With the imputed matrix  $\hat{\bX}_t$ at hand, the ground-truth image can then be reconstructed using the magnitude of $\cF^{-1}(\hat{\bX}_t)$, namely $[\hat{\bL}_t]_{i,j}=\big| [\cF^{-1}(\hat{\bX}_t)]_{i,j} \big|$.


The corresponding algorithm specialized to the MRI task is listed under Algorithm~\ref{tab:alg_interpolation_mri}. It can be seen as a special case of the general Algorithm~\ref{tab:alg_lowrank_sparse_decomp} upon fixing~$[\bPhi_t]_{\ell,r}=[\ba_r^{(1)}[t]]_{i} [\ba_r^{(2)}[t]]_{j}$ for $\ell \equiv (i,j)$. The iterations admit a simple and interpretable form, where the $i$ and $j$ rows of $\bA_1[t]$ and $\bA_2[t]$, respectively, are updated once the $k$-space datum $(i,j) \in \Omega_t$ arrives.

\begin{algorithm}[t]
	\caption{Online interpolation-based MRI} \small{
		\begin{algorithmic}
			\STATE \textbf{input} 
			$\{y_{t}^{(i,j)},~\ell \in \Omega_t\}_{t=1}^{\infty},\{\mu_t\}_{t=1}^{\infty},\lambda,R$.
			\STATE \textbf{initialize} $(\bA_1[0],\bA_2[0])$.
			\FOR {$t=1,2$,$\ldots$}
			\STATE \textbf{(S1) Projection coefficients update}                
			
			\STATE $[\bPhi_t]_{\ell,r}=[\ba_r^{(1)}[t-1]]_{i} [\ba_r^{(2)}[t-1]]_{j}$,~$\ell \equiv (i,j) \in \Omega_t$,~~~~$\bPhi_t=\bU_t \bSigma_t \bV_t^{\top}$
			\STATE $\bD_t={\rm diag}\big[\sigma_1(\sigma_1^2+\lambda)^{-1},\ldots,\sigma_R(\sigma_R^2+\lambda)^{-1} \big]$,
			
			\STATE $\bgamma_t=\bV_t \bSigma_t^{-1}\bD_t \bU_t^{\top} \by_t$

			\STATE $e_{t}^{(i,j)}:= y_{t}^{(i,j)}-\langle \bphi_{t}^{(i,j)}, \bgamma_t \rangle$

			\STATE \textbf{(S2) Parallel subspace update} $(r \in [R])$
			
			\STATE $\ba_r^{(1)}[t] = (1-\mu_t\lambda/t)\ba_r^{(1)}[t-1] + \mu_t \gamma_{t,r}^{*} \sum_{(i,j) \in \Omega_t}  e_{t}^{(i,j)}  (a_{j,r}^{(2)}[t-1])^{*} \be_{i}$
			
			\STATE $\ba_r^{(2)}[t] = (1-\mu_t\lambda/t)\ba_r^{(2)}[t-1]  
			+\mu_t \gamma_{t,r}^{*} \sum_{(i,j) \in \Omega_t}  e_{t}^{(i,j)}  (a_{i,r}^{(1)}[t-1])^{*} \be_{j}$

			\RETURN  $(\bA_1[t],\bA_2[t])$
			\ENDFOR
		\end{algorithmic}
	}
	\label{tab:alg_interpolation_mri}
\end{algorithm}

This interpolation-based approach is particularly attractive when one can split each $N_1 \times N_2$ $k$-space image into $K_1 \times K_2$ (non)overlapping patches of size $n_1 \times n_2$, with $K_1=N_1/n_1$ and $K_2=N_2/n_2$. Patching is known to better leverage the local image features~\cite{ravishankar2011mr}. Patches must be sufficiently sizable to preserve the spatio-temporal correlations, and form a tensor with low rank $\rho$. This idea reduces the variable count associated with the subspace from $(N_1+N_2)R$ to $(n_1+n_2)\rho$ which can lead to significant computational savings. In addition, the large number of frames facilitates learning the tensor subspace, especially for MRI scans with low temporal resolution. Moreover, subsampling strategies, discussed in Section \ref{sec:adapt_sampling}, are immediately applicable to reduce acquisition time. In the multi-coil acquisition scenario the coil sensitivity information can be further leveraged to improve the reconstruction quality. This issue is dealt with in the ensuing subsection.

\begin{figure}[t]
	\centering
	\includegraphics[scale=1.75]{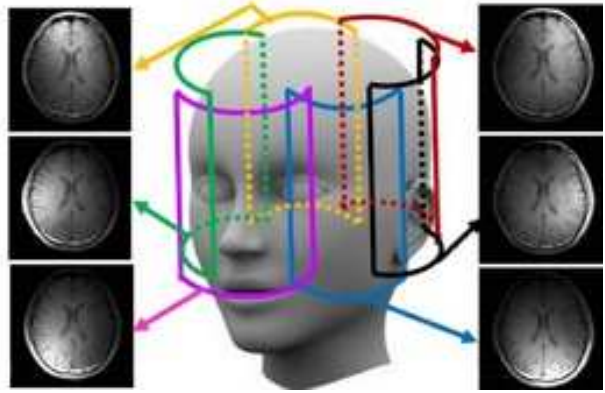}
	\caption{Multi-coil parallel MRI acquisition.}
	\label{fig:fig_singval}
\end{figure}

\subsection{Tomographic parallel MRI}
\label{subsec:tomo_par_mri}
While the approach of the previous subsection interpolates the missing $k$-space data and then inverts the $k$-space data to reconstruct the spatial domain image, one can instead pursue a tomographic approach to retrieve the images directly from partial $k$-space data. To generalize the measurement model \eqref{eq:mri_model}, we adopt the multi-coil scenario where the ground-truth image $\bL_t$ is acquired by $C$ coils, each sensitive to a specific region of the image. The $c$-th coil's sensitivity to the image pixels is modeled by the complex-valued matrix $\bH_c \in {\mathbb{C}}^{N_1 \times N_2}$. A large magnitude of $[\bH_c]_{i,j}$ indicates a `good view' of the $(i,j)$-th pixel. Ideally, the sensitivity matrices $\{\bH_c\}_{c=1}^C$ are expected to be non-overlapping, and cover the entire image, that is $\bH_c \odot \bH_{c'} = \mathbf{0}$, and $\sum_{c=1}^C \bH_c=\mathbbm{1}\mathbbm{1}^{\top}$. Let $\bL_t^{(c)}:=\bH_c \odot \bL_t$ denote the true image from the viewpoint of the $c$-th coil.

Suppose that the coil sensitivities $\{\bH_c\}_{c=1}^C$ have been estimated directly as in e.g.,~\cite{pruessmann1999sense}. With each coil $c$ acquiring the $k$-space data indexed by $\Omega_t$, our idea is to collect $C |\Omega_t|$ data per time instant $t$, where the $(i,j)$-th datum at coil $c$ adheres to
\begin{align}
y_c^{(i,j)}[t] = [\cF(\bH_c \odot \bL_t)]_{i,j} + v_t^{(i,j)},\quad (i,j) \in \Omega_t,~~c \in [C],~~t \in [T]. \label{eq:data_parallel_mri}
\end{align}
This tomographic data model in \eqref{eq:data_parallel_mri} is a special case of the general one in \eqref{eq:obs_model} when the subsampling matrix is given by
\begin{align}
\bW_c^{(i,j)}[t] = \bH_c \odot \cF ( \be_{i} \be_{j}^{\top} )
\end{align}
%

Various parallel imaging schemes have been introduced to combine the acquired images across coils for reconstructing the sought image. Among others, SENSE~\cite{pruessmann1999sense} and GRAPPA~\cite{griswold2002generalized} are commonly used in practice; see also~~\cite{deshmane2012parallel}. Each coil in the SENSE method, reconstructs an aliased image based on the subsampled $k$-space data (usually a fraction of phase encoding rows is selected). Then, the aliased images (or pixels) at various coils, each a linear combination of different pixels, are used to jointly reconstruct the ground-truth image~\cite{pruessmann1999sense}. Clearly, SENSE leverages the spatial diversity across coils. However, it requires knowledge of $\{\bH_c\}_{c=1}^C$. On the other hand, the GRAPPA technique works with the raw undersampled $k$-space data, and interpolates the missing samples from the present neighboring ones (through a Kernel that is obtained from a calibration process using additional $k$-space data), and subsequently reconstructs the image~\cite{griswold2002generalized}. The crux of GRAPPA is that the acquired $k$-space data per coil pertains to the ground-truth object weighted by the coil sensitivities, and thus the $k$-space of the object is convolved with the $k$-space of coil sensitivities, which smears the $k$-space information.

Our tomographic reconstruction approach relies on $\by_t$, which collects the complex-valued observations of coils $1$ to $C$ orderly on top of each other. Likewise, let $[\bPhi_t]_{\ell,r}:= [\cF(\bH_c \odot (\ba_r^{(1)}[t-1] \circ \ba_r^{(2)}[t-1]))]_{i_{c},j_{c}}$ be a complex regression matrix comprising $C |\Omega_t|$ columns, where columns $(c'-1) |\Omega_t| +1$ till $c' |\Omega_t|$ correspond to coil $c'$. With $e_{c}^{(i,j)}$ denoting the fitting error for the $(i,j)$-th datum at coil $c$, due to linearity of the Hadamard product one can simplify the gradient update using the fact that
\begin{align}
\boldsymbol{\Theta}[t]:=\sum_{c=1}^C \sum_{(i,j) \in \Omega_t} e_{c}^{(i,j)}[t] (\bW_{c}^{(i,j)}[t])^* =\sum_{c=1}^C \bH_c^{*} \odot \cF^{-1}(\boldsymbol{\Xi}_c[t])
\end{align}
where $[\boldsymbol{\Xi}_c[t]]_{i,j} := e_{c}^{(i,j)}[t],~(i,j) \in \Omega_t$, and zero otherwise. 

Following the general steps in Algorithm~\ref{tab:alg_lowrank_sparse_decomp}, the iterations of the novel parallel MRI scheme with  estimated coil sensitivities, are listed under Algorithm~\ref{tab:table_sensitivity_par_mri}.


\begin{algorithm}[t]
\caption{Online sensitivity-aware and tomographic parallel MRI} \small{
\begin{algorithmic}
	\STATE \textbf{input} 
	$\{y_{c}^{(i,j)}[t],~(i,j) \in \Omega_t,~~c=1,\ldots,C\}_{t=1}^{\infty},\{\mu_t\}_{t=1}^{\infty},\lambda,R,C$.
    \STATE \textbf{initialize} $\{\ba_r^{(1)}[0],\ba_r^{(2)}[0]\}_{r=1}^R$ at random.
    \FOR {$t=1,2$,$\ldots$}
    
                \STATE \textbf{(S1)~Principal components update}
                \STATE $[\bphi_{\ell}^{(c)}]_{r}=\Big[ \cF \big( \bH_c \odot (\ba_r^{(1)}[t-1] \circ \ba_r^{(2)}[t-1]) \big) \Big]_{i,j},~~\ell \equiv (i,j)$ 
                
                \STATE $\bPhi_t = [\bphi_{1}^{(1)}, \ldots,\bphi_{|\Omega_t|}^{(1)},\ldots, \bphi_{1}^{(C)},\ldots,\bphi_{|\Omega_t|}^{(C)}]^{\top}, \quad \bPhi_t=\bU_t \bSigma_t \bV_t^{\top}$
                \STATE $\bD_t={\rm diag}\big[\sigma_1(\sigma_1^2+\lambda)^{-1},\ldots,\sigma_R(\sigma_R^2+\lambda)^{-1} \big]$
                \STATE $\bgamma_t=\bV_t \bD_t \bU_t^{\top} \by_t$
                \STATE $e_{c}^{(i,j)}[t]:= y_{c}^{(i,j)}[t] - \langle \bphi_{\ell}^{(c)}, \bgamma_t \rangle$
                
                \STATE \textbf{(S2)~Parallel subspace update}~$\big(r \in [R]\big)$
                
                \STATE $\boldsymbol{\Theta}[t]  = \sum_{c=1}^C \bH_c^* \odot \cF^{-1}(\boldsymbol{\Xi}_c[t])$
                
                 \STATE $\ba_r^{(1)}[t]=(1-\lambda \mu_t/t)\ba_r^{(1)}[t-1]  + \mu_t\gamma_{t,r}^* \boldsymbol{\Theta}[t]  (\ba_r^{(2)}[t-1])^*  $
                 
                \STATE $\ba_r^{(2)}[t]=(1-\lambda \mu_t/t)\ba_r^{(2)}[t-1]  + \mu_t \gamma_{t,r}^* \boldsymbol{\Theta}^{\top}[t]  (\ba_r^{(1)}[t-1])^* $

                \RETURN  $\{\ba_r^{(1)}[t],\ba_r^{(2)}[t]\}_{r=1}^R$
    \ENDFOR
\end{algorithmic}}
\label{tab:table_sensitivity_par_mri}
\end{algorithm}

\section{Subspace Learning via Randomized Subsampling}
\label{sec:adapt_sampling}
While the missing data is often attributed to outliers or malfunctioning of the acquisition process, one can purposely skip data to either facilitate the acquisition process, or lower the computational burden. This well aligned with recent efforts towards accelerating the long MRI scans, which create artifacts especially when imaging moving objects. Imagine for instance the MR scanner knowing {\rm a priori} the best minimal subset of $k$-space data to collect per cardiac snapshot. It has then sufficient time to acquire important samples before the heart moves to a new state.

In essence, design of the optimal subsampling needs knowledge of the underlying unseen physical phenomenon of interest, that is practically infeasible. Typical sketching strategies, e.g., in the context of matrix sparsification, assume data are {\it fully} available to score, and subsequently select a subset of them; see e.g.,~\cite{mahoney2011randomized}. However, in MRI applications, {\it data acquisition} is the main challenge~\cite{lauterbur1973image}, and data are streaming as well. All in all, given the subspace estimates offered by the online iterates until time $t-1$, our goal is to {\rm adaptively} design/predict the sketching operator $\{\ubW_{t}^{(\ell)}\}_{\ell=1}^{L_t}$ that results in a minimal preselected sample count $L_t$, while attaining a prescribed reconstruction quality. In what follows, we focus on subsampling as a special sketching operator that picks only a subset of tensor entries per time instant. This subset is denoted by $\Omega_t \subset [N_1] \times \ldots \times [N_M]$. The overall learning task in (P2) then amounts to tensor imputation.

\subsection{Importance scores}
\label{subsec:imp_scores}
Albeit streaming data poses an extra challenge to sketching (since future observations are not available), online learning offers intermediate estimates of the latent tensor subspace, namely $\{\bA_i[t-1]\}_{i=1}^{M-1}$, that can be leveraged to devise adaptive subsampling strategies. In order to predict $\Omega_t$, our basic idea is to rank the samples (entries) according to their level of importance measured by a certain score along the lines of~\cite{mahoney2011randomized} and \cite{chen2014coherent}. However, different from the online setup  dealt with here,~\cite{mahoney2011randomized} and \cite{chen2014coherent} assume batch data processing.

To gain insight, it is instructive to start with the three-way array ($M=3$) having the entire tensor data $\ubY$ at hand, and seeking the factor matrices $\{\bA_1,\bA_2\}$. The $(n_1,n_2)$-th entry of $t$-th tensor slice can then be expressed as 
\begin{align}
[\bY_t]_{n_1,n_2} \approx \sum_{r=1}^R \gamma_{t,r} a_{n_1,r}^{(1)} a_{n_2,r}^{(2)}  \nonumber
\end{align}
where $\{\gamma_{t,r}\}_{r=1}^R$ are shared by every sample in the same slice $t$. The samples are distinguished through their weight vectors $\{a_{n_1,r}^{(1)}\}_{r=1}^R$ and $\{a_{n_2,r}^{(2)}\}_{r=1}^R$, corresponding to $n_1$- and $n_2$-th rows of $\bA_1$ and $\bA_2$, respectively. Broadening the scope of \cite{mahoney2011randomized,chen2014coherent} to three-way arrays with PARAFAC decomposition, our proposed metric to score the $(n_1,n_2)$-th feature is based on the energy of the corresponding rows in the subspace matrices $\bA_1$ and $\bA_2$.

According to \cite{chen2014coherent} the informative samples are the ones that if missed, the matrix cannot be recovered reliably. Consider the $t$-th tensor slice $\bY_t$ with the SVD $\bU\bSigma\bV^{\top}$. Matrix completion literature captures the information content of $(i,j)$-th entry through the so-termed {\it local coherence} measure $s_i^u:=\|\bU^{\top}\be_i\|^2$ and $s_j^v:=\|\bV^{\top}\be_j\|^2$ associated with $i$-th row and $j$-th column, respectively. In essence, the $(i,j)$ entry is informative when both $s_i^u$ and $s_j^v$ are large, i.e., have large projection onto the column and row space of the matrix $\bY_t$. In particular, \cite{chen2014coherent} adopts the score $s_i^u + s_j^v$ to rank entries based on their level of importance. To gain further intuition, imagine the perturbed matrix $\bY_t+\bH$ with $(i,j)$-th entry perturbed by $\bH=h\be_i\be_j^{\top}$ for some $h$. A large $s_i^u$ and $s_j^v$ implies that $\bH$ is well spanned by the column and row space of $\bY_t$, and as a result $\bY_t$ and $\bY_t+\bH$ have the same rank. Hence, $\bY_t$ can be misidentified as $\bY_t+\bH$; see e.g., \cite{chen2014coherent} for more technical details. Note also that in the matrix sparsification context~\cite{mahoney2011randomized} the coherence measure is referred to as the statistical leverage score which has direct ties to the residual variance of the LS estimation~\cite{mahoney2011randomized}. For a three-way array, the column vectors $\{\ba_r^{(1)}\}_{r=1}^R$~(resp. $\{\ba_r^{(2)}\}_{r=1}^R$) of $\bA_1$ (resp. $\bA_2$) span the column- (resp. row-) space of the $t$-th tensor slice. Albeit not necessarily orthonormal, $\bA_1$ and $\bA_2$ play a role similar to the orthonormal factors $\bU$ and $\bV$

%
%

Along this line of thought, consider now the online setup where at time instant $t$ one has access to the subspace estimate $(\bA_1[t-1],\bA_2[t-1])$, and aims to acquire a few samples from the next slice $\bY_t$ indexed by $\Omega_t$. Suppose also that slices $\{\bY_t\}$ change slowly over time; this is the case for instance in dynamic cardiac MRI where different slices correspond to different snapshots of a patient's beating heart. Assume further that $(\bA_1[t-1],\bA_2[t-1])$ provide a {\it reliable} estimate of the underlying tensor subspace e.g., as a result of a warm initialization. It is then reasonable to adopt the metric $\|\bA_1^{\top}[t-1]\be_{n_1}\|_F^2 + \|\bA_2^{\top}[t-1]\be_{n_2}\|_F^2$ to predict the information content of the $(n_1,n_2)$-th feature at time $t$.

Apparently, if $\bY_t$ contains innovation, not captured by the subspace $(\bA_1[t-1],\bA_2[t-1])$ that is learned from past data $\{\bY_{\tau}\}_{\tau=1}^{t-1}$, the informative samples may be misidentified. To cope with this issue, and avoid sampling bias due to initialization, our proposed subsampling strategy is randomized as discussed next.

\subsection{Randomized sketching}
\label{subsec:prob_sampling}
Normalize columns of $\bA_1[t-1]$ and $\bA_2[t-1]$ to end up with $\bar{\bA}_1[t-1]$ and $\bar{\bA}_2[t-1]$, respectively. For notational brevity, let also~$\bar{\bA}_1[t-1]:=[\bar{\balpha}_1^{(1)},\ldots,\bar{\balpha}_{N_1}^{(1)}]^{\top}$, and~$\bar{\bA}_2[t-1]:=[\bar{\balpha}_1^{(2)},\ldots,\bar{\balpha}_{N_2}^{(2)}]^{\top}$, and score the sample $(n_1,n_2)  \in [N_1] \times [N_2]$ at time $t$ using
\begin{align}
s_t(n_1,n_2):=\frac{1}{R(N_1+N_2)}\Big( \big\|\bar{\balpha}_{n_1}^{(1)} \big\|^2 + \big\|\bar{\balpha}_{n_2}^{(2)} \big\|^2 \Big).  \label{eq:score}
\end{align}
The scores $\{s_t(n_1,n_2)\}$ are positive valued and sum up to unity; thus, one can interpret them as a probability distribution over the entries. For a prescribed maximum sample count $K$, one can then draw $K$ random trials from the distribution $s_t$ to collect the important samples in the set $\Omega_t$. 

Sampling can be performed with, or, without replacement depending on how the image energy is distributed. Sampling without replacement results in exactly $K$ samples. However, sampling with replacement can even reduce the sample count ($|\Omega_t|<K$) when the samples are nonuniformly informative. For MRI images typically most of the energy is concentrated in low-frequency $k$-space samples. Thus, it seems more natural to consider random draws with replacement which is more likely to discard the less informative samples (with tiny score) that have only a negligible contribution to the image. It is also worth commenting that nowadays MRI scanners can accommodate quickly changing the gradient pulse sequence to acquire the sampled phase encoding lines in real time~\cite{zhi2000principles}.

Adopting randomized sketching along with the online iterates for subspace learning in \eqref{eq:gradeint_iteration}, the resulting procedure for a general $M$-way data array is listed under Algorithm~\ref{tab:alg_rnd_sampling}. The iterative scheme begins with a warm initialization, obtained for instance after first running the algorithm over a small training dataset. Each iteration (time instant) $t$ comprises three major steps, where the first step (S0) {\it probabilistically} decides on the subsampling set $\Omega_t$, that is subsequently used to acquire the corresponding samples in $\bY_t$. Based on the partial samples in $\Omega_t$, the second step (S1) finds the principal components of $t$-th frame across the subspace bases $\{\ba_r^{(1)}[t-1]{\ba_r^{(2)}}^{\top}[t-1]\}_{r=1}^R$. The innovation of the new (imputed) datum captured through the error term $\{e_t^{(i,j)} \}_{(i,j)\in\Omega_t}$, in the third step (S2) then refines the subspace bases.

One important question at this point pertains to the (average) number of samples acquired per slice by the random subsampling with replacement. This depends on the degree of sample  nonuniformity, and the initialization. In an extreme case with equally important entries, exactly $K$ samples are acquired per time slot, which can be considerably lower for nonuniform entries. To see this, introduce a random variable $X_{m,n}$ denoting the frequency of choosing the $(n_1,n_2)$-th entry after $K$ trials. The random sample count is then $|\Omega_t|=\sum_{n_1,n_2} \mathbbm{1}_{\{X_{n_1,n_2}\}}$, where the indicator $\mathbbm{1}_{\{x \}}$ takes value of one if $x>0$, and zero otherwise. In general, the random variables $\mathbbm{1}_{\{X_{n_1,n_2} \}}$ are dependent, which renders the distribution analysis for $|\Omega_t|$ formidable. The expected sample count per slice however can be expressed as
\begin{align}
\mathbb{E}[|\Omega_t|]=\sum_{m=1}^M \sum_{n=1}^N \Big(1-[1-s_t(n_1,n_2)]^K \Big). \nonumber 
\end{align} 
Finally, the average sample count across time and entries is given by $\bar{N}_t:=(1/t)\sum_{\tau=1}^{t} \mathbb{E}[|\Omega_{\tau}|]$.



\begin{algorithm}[t]
	\caption{Random subsampling for $M$-way arrays} \small{
		\begin{algorithmic}
			
			\STATE \textbf{input} $\{\bA_m[t-1]\}_{m=1}^{M-1}$, $K$, and $R$
			
			\vspace{1mm}
			
			\STATE $\bar{\bA}_m[t-1]:=\bA_m[t-1] {\rm diag}\big(\|\ba_1^{(m)}[t-1]\|,\ldots,\|\ba_R^{(m)}[t-1]\| \big)^{-1}$ 
			
			
			
			\vspace{1mm}
			
			\STATE $\bar{\bA}_m[t-1]:=[\bar{\balpha}_1^{(m)},\ldots,\bar{\balpha}_{N_m}^{(m)}]^{\top}$ 
			
			\vspace{1mm}
			
			\STATE $s_t(n_1,\ldots,n_{M-1}):=\frac{1}{R \sum_{m=1}^{M-1} N_m} \sum_{m=1}^{M-1} \|\bar{\balpha}_{n_m}^{(m)}\|^2 $
			
			\vspace{1mm}
			
			\STATE Draw $K$ random trials from $[N_1] \times \ldots \times [N_M]$ based on $s_t$ to form $\Omega_t$
			
			\vspace{1mm}
			
			
			\vspace{1mm}  
			
			\STATE \textbf{output}  $\Omega_t$
			
		\end{algorithmic}}
		\label{tab:alg_rndom_acquisition}
	\end{algorithm}

	\begin{algorithm}[t]
		\caption{Randomized tensor subspace learning for imputation of $M$-way arrays} \small{
			\begin{algorithmic}
				
				\STATE \textbf{input} $\{\mu_t\}_{t=1}^{\infty}, K, R, \lambda$
				
				\vspace{1mm}
				
				\STATE \textbf{initialize} $\{\bA_m[0]\}_{m=1}^{M-1}$ with a warm startup
				
				\vspace{1mm}
				
				\FOR {$t=1,\ldots $ }
				
				\vspace{1mm}
				
				\STATE \textbf{(S0) Random subsampling}
				
				\vspace{1mm}
				
				\STATE Acquire $\{y_t^{(n_1,\ldots,n_{M-1})}\}_{(n_1,\ldots,n_{M-1}) \in \Omega_t}$ based on Algorithm 1  
				
				\vspace{1mm}  
				
				\STATE \textbf{(S1) Principal components update}
				
				\vspace{1mm}
				
				\STATE $[\bPhi_t]_{(n_1,\ldots,n_{M-1}),r}=\prod_{m=1}^{M-1}a_{n_m,r}[t-1]$ ,~~$\bPhi_t=\bU_t \bSigma_t \bV_t^{\top}$

				\STATE $\bD_t={\rm diag}\big[\sigma_1(\sigma_1^2+\lambda)^{-1},\ldots,\sigma_R(\sigma_R^2+\lambda)^{-1} \big]$,
				
				\STATE $\bgamma_t=\bV_t \bSigma_t^{-1}\bD_t \bU_t^{\top} \by_t$

				\STATE $e_{t}^{(n_1,\ldots,n_{M-1})}:= y_{t}^{(n_1,\ldots,n_{M-1})}-\langle \bphi_{t}^{(n_1,\ldots,n_{M-1})}, \bgamma_t \rangle$
				
				\vspace{1mm}
				
				\STATE \textbf{(S2) Parallel subspace update} $[(m,r) \in [M] \times [R]]$
				
				\STATE $\ba_r^{(m)}[t] = (1-\mu_t\lambda/t)\ba_r^{(m)}[t-1] $
				
				\STATE $\hspace{2cm} +\mu_t \gamma_{t,r}^{*} \sum_{(n_1,\ldots,n_M) \in \Omega_t}  e_{t}^{(n_1,\ldots,n_{M-1})} \prod_{m=1,m \neq m}^{M-1} (a_{n_m,r}^{(m)}[t-1])^{*} \be_{m}$
				
				
				
				\vspace{1mm}

				\ENDFOR
				\STATE \textbf{output}  $\{\bA_1[t],\bA_2[t]\}$
				
			\end{algorithmic}}
			\label{tab:alg_rnd_sampling}
		\end{algorithm}

\section{Numerical Tests}
\label{sec:sim}
Performance of the novel tensor subspace learning scheme for online MRI reconstruction is assessed in this section using real cardiac MRI data. Two datasets are adopted as follows: (D1) single-coil acquisition of $256$ cardiac cine MRI frames, each of size $200 \times 256$ pixels; and, (D2) $16$-coil dataset including $26$ frames of size $192 \times 120$ acquired by the Center for Magnetic Resonance Research (CMRR) at the University of Minnesota. The experiments were run using Matlab R2016a on a Ubuntu 15.10 machine with 64GB RAM. The processors include an Intel core i7 CPU, and an NVIDIA GDX 970 GPU. 

For validation purposes the $k$-space data is fully acquired. To simulate the real-world undresampled data, as it is common with practical MRI scanners, variable density Cartesian sampling is used to randomly sample a small fraction $\pi$ of the phase encoding lines. They are sampled based on the polynomial distribution $p_i=i^{\alpha}/\sum_{i=1}^{M/2-1} i^{\alpha}$, where $i$ is the distance from the center line. The center line ($i=0$) is always chosen, and we fix $\alpha=-1$, so that the low frequency components carrying most of the image energy are more likely to be selected.

\subsection{Real-time reconstruction}
\label{subsec:real_time_recon}
The frequency-domain interpolation-based scheme in Algorithm~\ref{tab:alg_interpolation_mri} is tested on the dataset (D1) forming the tensor $\ubX \in \mathbbm{C}^{200 \times 256 \times 256}$. The singular values of the unfolded tensor along the temporal axis, as well as the $k_x$ and $k_y$ axes are plotted in Fig.~\ref{fig:fig_singval}. The unfolded $k$-space data matrix along the temporal dimension exhibits only about four dominant singular values, indicating high spatio-temporal correlations indicative of low rank for tensor $\ubX$. The first $5$ scans are fully acquired, and used as training data to provide a warm startup for the tensor subspace initialization. In practice, such a warm initialization can be even provided by historical data of other patients, or, a complementary prescan of the same patient. For the rest of $251$ frames, partial $k$-space data are acquired based on the variable density Cartesian sampling.



\begin{figure}[t]
\centering
\begin{tabular}{ccccc}
     \hspace{-10mm}\epsfig{file=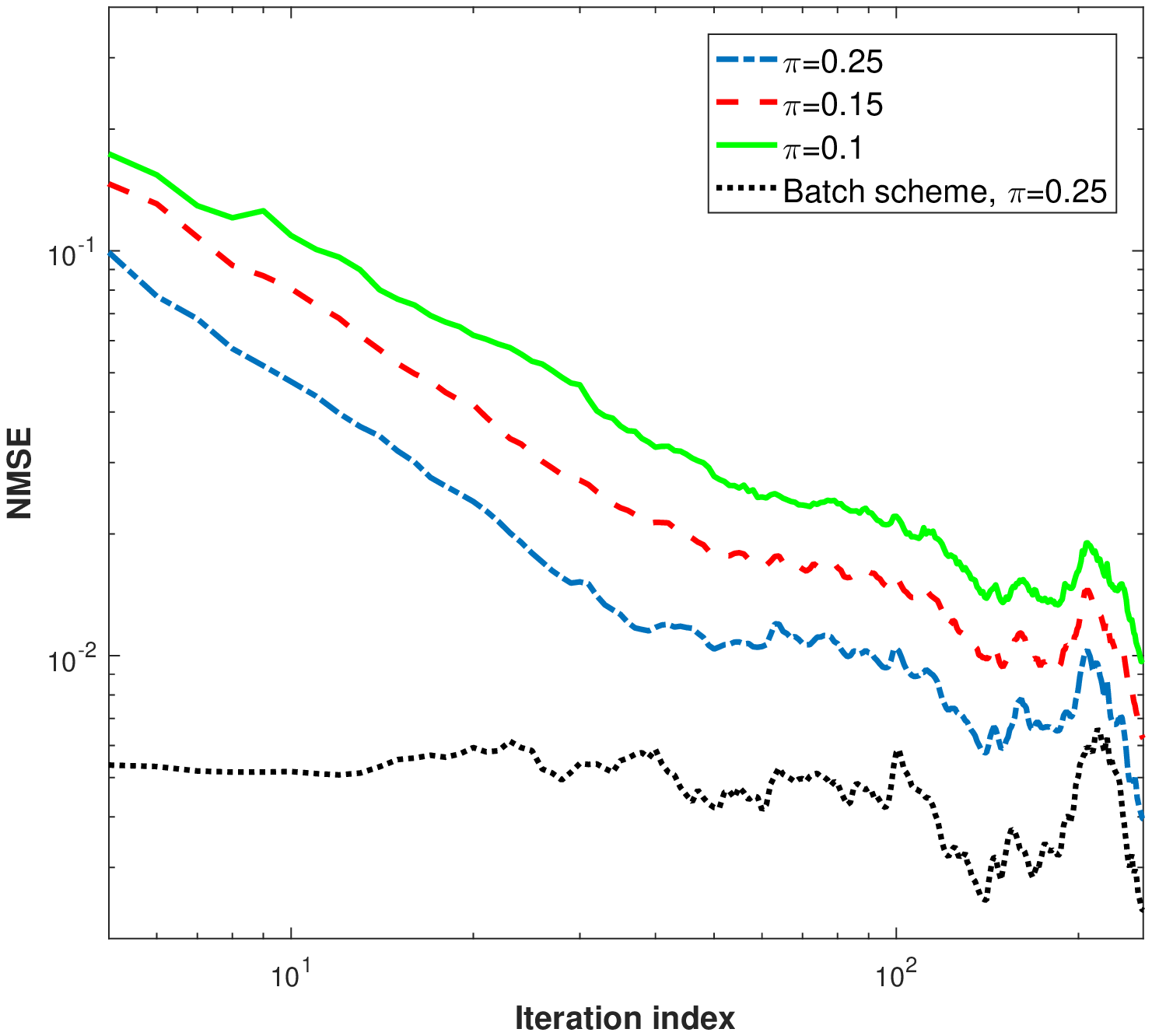,width=0.5
     \linewidth, height=2.75 in } &
     \hspace{-8mm}\epsfig{file=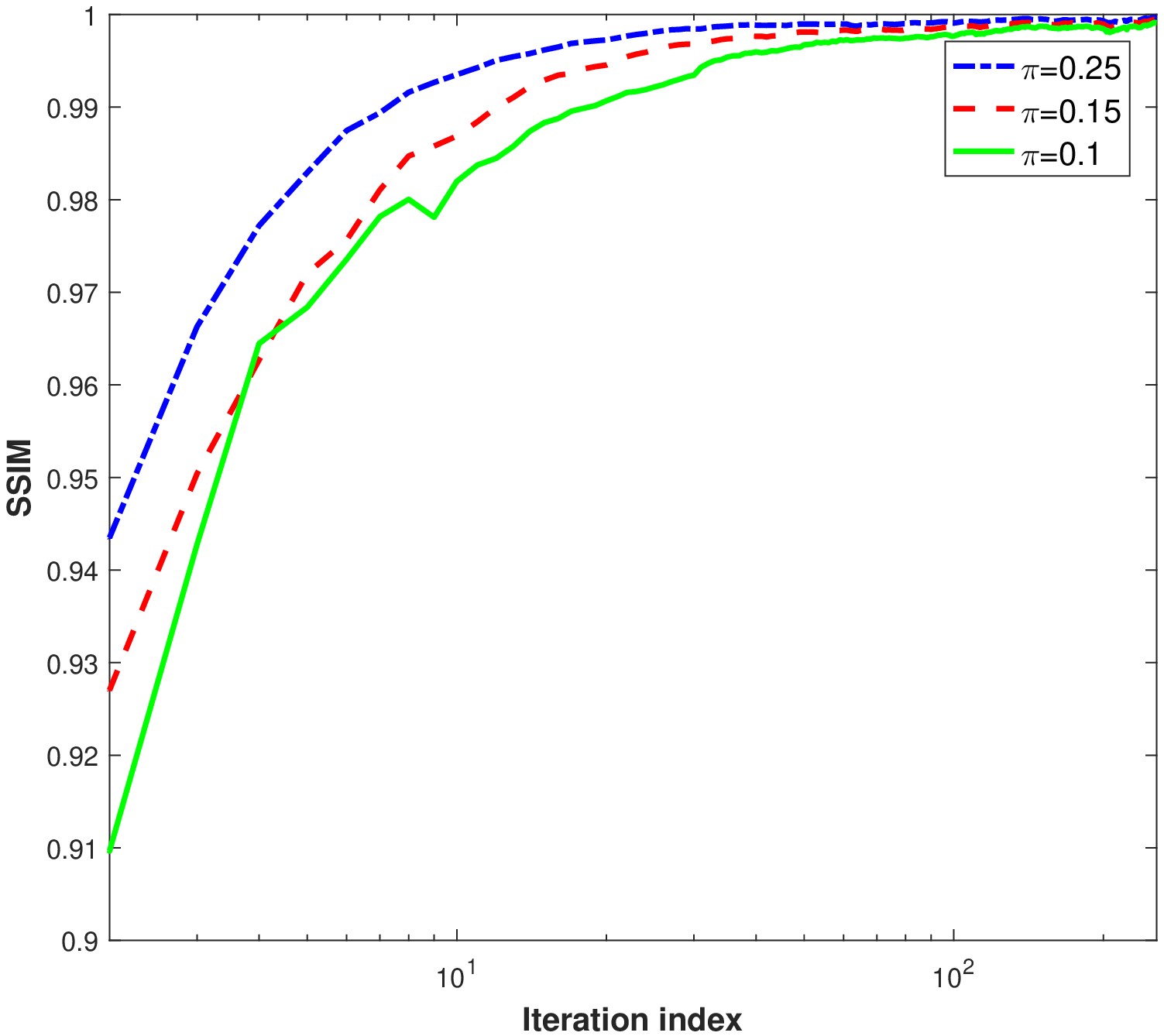,width=0.5
     \linewidth, height=2.75 in } \\
     (a) &
     (b)
  \end{tabular}
  \caption{Evolution of NMSE (left) and SSIM (right) over iteration index (time) when a fraction $\pi$ of phase encoding lines are sampled ($R=100$, $\lambda=2$, $\mu=0.01$). }
  \label{fig:fig_cardiac_mri}
\end{figure}
\begin{figure}[t]
	\hspace{-1.7cm}\includegraphics[scale=1.05]{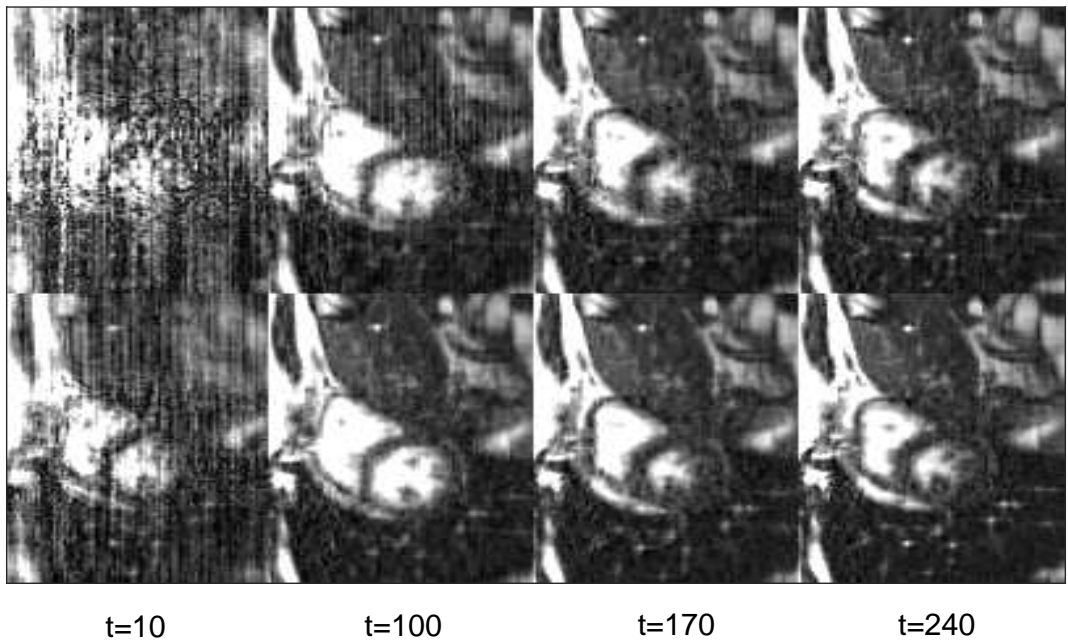}
  \caption{Real-time reconstruction of {\it in vivo} MRI dataset based on Algorithm~\ref{tab:alg_interpolation_mri} at different time points $t=10,100,170,240$ (left to right) under (top) $10$-fold, and (bottom) $4$-fold acceleration, when $R=100$. A patch of size $138 \times 100$ pixels is shown.}
  \label{fig:fig_cardiac_mri_realtime_interpolation}
\end{figure}

Upon choosing the step size $\mu_t=0.01$ and $\lambda=2$, we plot time evolution of the normalized mean-square error ${\rm NMSE}_t=\|\bX_t-\hat{\bX}_t\|_F^2 / \|\bX_t\|_F^2$ with $\bX_t$ and $\hat{\bX}_t$ denoting, respectively, the true and estimated frames per $t$-th frame. As a subjective metric, structural similarity index (SSIM)~\cite{wang2004image} is also plotted over time in Fig.~\ref{fig:fig_cardiac_mri} under different tensor ranks and acceleration rates. The resulting reconstructed images at $t=10,100,170,240$ are also shown and compared against the gold standard (full $k$-space data) in Fig.~\ref{fig:fig_cardiac_mri_realtime_interpolation}. Apparently, the first $50$ frames show a sharp improvement in learning the subspace and consequently the reconstruction accuracy, which improves gradually for the subsequent frames. More epochs over the data (compared to the single visit for real-time scheme) are needed to fully learn the subspace. It turns out that $5$ epochs suffice to learn the subspace, where the output of each epoch serves to initialize the subspace for the next one. The reconstruction error for this offline scheme is also shown in Fig.~\ref{fig:fig_cardiac_mri} as the benchmark. Note that under $10$-fold acceleration, the acquisition time for each frame is about xxxx sec while for $R=100$ the reconstruction takes about $0.1$ sec on a GPU. This can be further shorten to $0.05$ sec with $R=50$ that yields relatively close reconstruction quality; NMSE=$0.029,0.03$ for $R=100,50$, respectively. Therefore, the novel scheme can reconstruct $20$ frames per second, which in turn makes {\it real-time} reconstruction practically feasible.

\begin{figure}[t]
\centering
\includegraphics[scale=0.6]{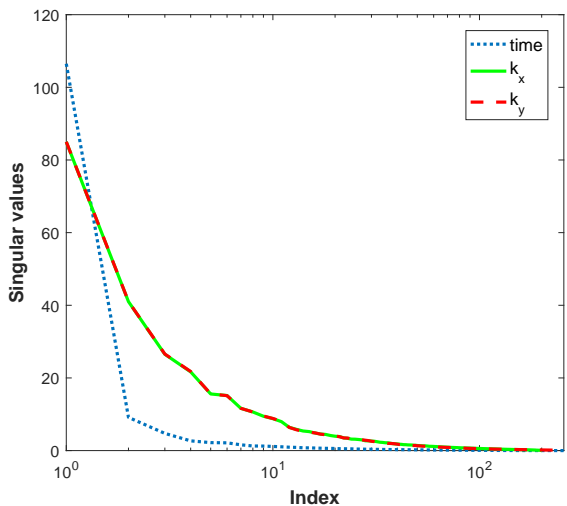}
\caption{Singular values of the matrix each corresponding to the unfolded $k$-space tensor across a different dimension.} 
\label{fig:fig_singval}
\end{figure}

%

\begin{figure}[t]
	\centering
	\begin{tabular}{cc}
		\epsfig{file=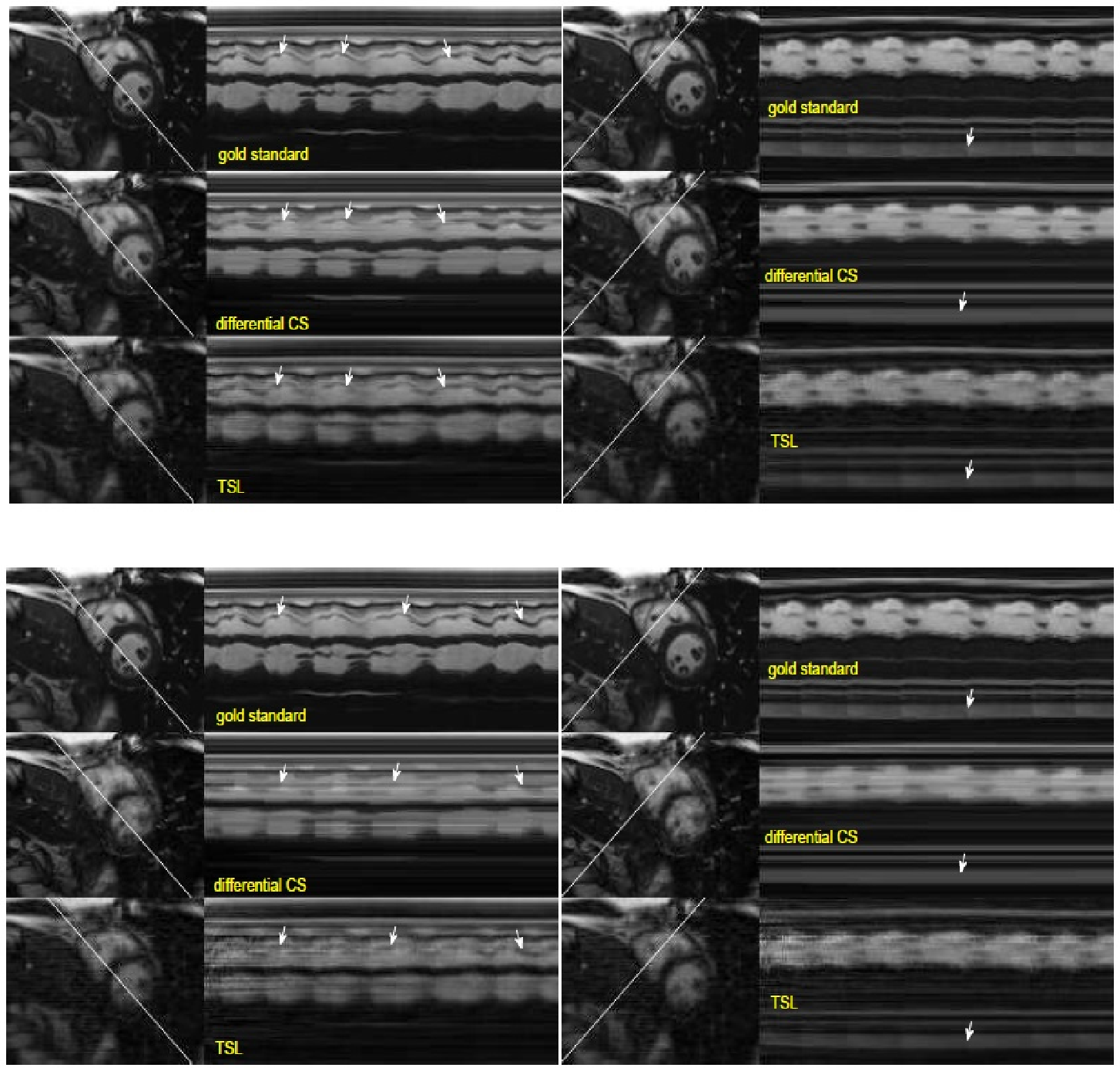,width=0.95 \linewidth, height=4.75 in } 
	\end{tabular} \vspace{0cm}
	\caption{Temporal profile of the oblique lines on the left side for the gold standard, differnetial CS, and the proposed TSL method for $4$-fold (top) and $10$-fold (bottom) acceleration under variable density sampling when $R=100$ and only the first five frames are used for training. The pixels on the lines are arranged vertically in the profile images.}
	\label{fig:fig_motion_map}
\end{figure}

%

\subsection{Comparisons}
\label{subsec:scheme_comparison}
Online MRI approaches so far include the Kalman filtering based ones~\cite{majumdar2016impulse,sumbul2009practical}, and the more recent compressive sampling (CS) based versions e.g., in~\cite{vaswani2010ls,yu2014multidimensional,majumdar2012compressed}. The Kalman filtering method enjoys relatively fast reconstruction at the expense of low image quality that CS schemes improve upon, at the expense of slower reconstruction that is due to the non-smooth regularization. Indeed, existing online reconstruction schemes are not real time since the reconstruction time is typically slower than the acquisition time.

To evaluate the merits of our novel reconstruction scheme, we compare with the differential CS scheme~\cite{majumdar2012compressed} as an state-of-the-art alternative. The crux of~\cite{majumdar2012compressed} is that the difference between subsequent frames is sparse. Let $\bX_{t-1}$ be an estimate of the previous frame $t-1$. To reconstruct $\bX_t$ the difference frame $\tilde{\bX}:=\bX_t-\bX_{t-1}$ is assumed sparse, and can thus be recovered by solving the non-smooth LASSO~\cite{tibshirani1996regression} program
\begin{align}
\tilde{\bX}_t = \arg\min_{\tilde{\bX}}~\sum_{(i,j) \in \Omega_t} \big(y_t^{(i,j)} - [\cF(\bX_{t-1})]_{i,j} - [\cF(\tilde{\bX})]_{i,j}\big)^2 + \lambda \|\tilde{\bX}\|_1. \label{eq:lasso}
\end{align}
Regarding the differential CS, it is apparently slow as it demands solving a LASSO program per time instant. Additionally, the error can gradually accumulate over time. SpaRSA package~\cite{wright2009sparse} is used to return the LASSO solution in~\eqref{eq:lasso} where for the stop criterion we choose the duality gap to be less than $0.01$. Each iteration is initialized with the difference image obtained in the previous iteration as a warm start up, and the maximum number of iterations is confined to $100$.

Temporal profile of the reconstructed images along the oblique lines are shown in Fig.~\ref{fig:fig_motion_map} for $4$-fold and $10$-fold accelerations with variable density Cartesian sampling. Parameter choices were $\lambda=0.001$ for differential CS, and $\mu=0.01$, $\lambda=2$, and $R=100$ for our novel method. As pinpointed by the arrows, differential CS leads to temporal blurring artifacts, while the proposed scheme can track the changes more accurately, and thus reveals the detailed temporal changes. Table~\ref{tab:table_runtime_tsl_diffcs} also lists the lower reconstruction time for TSL scheme relative to the differential CS scheme. This becomes more pronounced for higher acceleration rates as the operation count $\mathcal{O}(|\Omega_t| R^2)$ decreases (c.f. Remark 1), while the degree of non-smoothness for the LASSO program \eqref{eq:lasso}, and consequently the convergence time, increases. Note that more than $90\%$ of the TSL runtime per iteration stems from solving (exactly) the ridge-regression task (S1) to update $\bgamma_t$, which can be further reduced if one resorts to an inexact solution.

\begin{table}[t]
	\caption{Average frame reconstruction time (seconds) for TSL and differential CS schemes.}
	\vspace{-2.5mm}
	\label{tab:table_runtime_tsl_diffcs}
	\begin{center}
		\begin{tabular} {|c|c|c|c|c|}
			\hline
			Acceleration &  Diff. CS & TSL, $R=100$ (CPU) & TSL, $R=100$ (GPU) & TSL, $R=50$ (GPU)  \\
			\hline\hline
			$10$-fold & $0.55$ & $0.14$ & $0.11$ & $0.05$ \\
			\hline
			$4$-fold & $0.56$ & $0.28$ & $0.23$ & $0.1$ \\
			\hline
		\end{tabular}
	\end{center}
\end{table}

%






\subsection{Random $k$-space subsampling}
\label{subsec:adaptive_sketching}
\begin{figure}[t]
	\centering
	\includegraphics[scale=0.5]{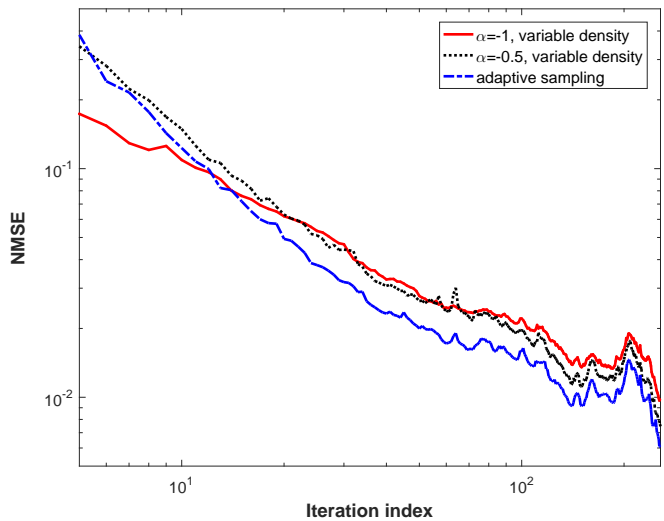}
	\caption{NMSE of the proposed score-based adaptive phase encoding versus the variable density counterpart with $10$-fold acceleration and $R=100$. } 
	\label{fig:fig_nmmse_adaptivesampling}
\end{figure}

\begin{figure}[t]
	\centering
	\includegraphics[scale=0.7]{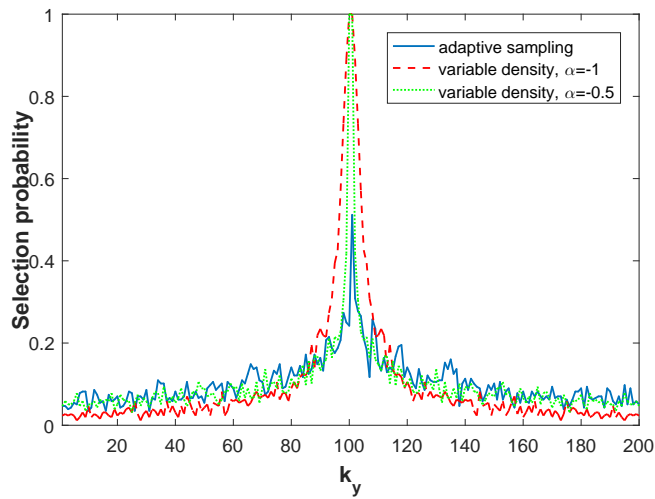}
	\caption{Sampling distribution $\rho$ for data driven (adaptive) and variable density schemes when $\pi=0.1$.} 
	\label{fig:fig_selectionrate_adaptivesampling}
\end{figure}

\begin{figure}[t]
	\centering
	\includegraphics[scale=1.5]{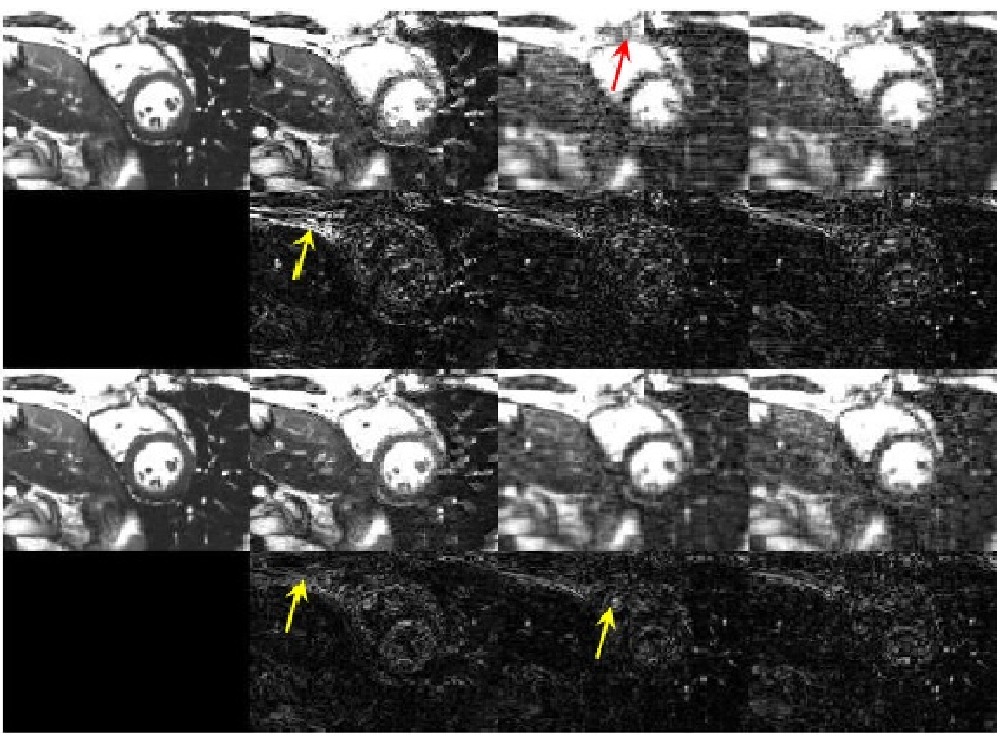} 
	\caption{Real-time reconstruction comparison at time instant $t=100$ for differential CS and TSL schemes under variable density and adaptive sampling. Top (bottom) row corresponds to $10$-fold ($4$-fold) acceleration. first column: original frame; second column: differential CS; third column: TSL with variable density sampling; and fourth column: TSL with adaptive sampling.}
	\label{fig:fig_cardiac_mri_realtime_comparsion}
\end{figure}

The random subsampling policy developed in Section~\ref{sec:adapt_sampling} is tested here to sample Cartesian phase encoded lines for interpolation of missing $k$-space data in dataset (D1). Since only the rows are selected, the score in~\eqref{eq:score} is marginalized over the columns to arrive at the modified score $s_t(n_1)=\frac{1}{R(N_1+N_2)} (N_2\|\bar{\balpha}_{n_1}^{(1)}\|^2+R)$, and subsequently random trials are drawn with replacement to sample the rows. As before, the first five frames are fully acquired to serve as a warm subspace startup. Evolution of NMSE is depicted in Fig.~\ref{fig:fig_nmmse_adaptivesampling} under $10$-fold acceleration for the adaptive sampling scheme, and its non-adaptive counterpart using variable density sampling with parameters $\alpha=-1,-0.5$ for the polynomial distribution $p_i$. After dozen iterations the latent structure in $k$-space data is gradually learned and as a result subspace-driven sampling starts outperforming the variable density sampling. This observation suggests that one better adopt variable density sampling with $\alpha=-1$ to pick the low frequency components for the early iterations to end up with a reliable subspace estimation, and then switch to adaptive subsampling for improved quality. The reconstructed frames at time $t=100$ are shown in Fig.~\ref{fig:fig_cardiac_mri_realtime_comparsion}, where one can confirm from the residual images that our novel scheme creates less artifacts as delineated by the arrows. For the resultant sampling patterns with $256$ temporal realizations, the sampling distribution for phase encoded lines is also depicted in Fig.~\ref{fig:fig_selectionrate_adaptivesampling}.

\subsection{Batch MRI processing}
\label{subsec:batch_mri}
The recursive structure and the lightweight iterations offered by the TSL scheme suit batch processing of large MRI datasets. Inspired by the incremental gradient methods, one can allow multiple epochs over the data, where the subspace learnt in the first epoch initializes the second epoch (with possibly random re-ordering of the frames), and so on. Batch performance of TSL is compared against the {\it k-t} FOCUSS scheme~\cite{jung2009k} that achieves the reconstruction quality of {\it k-t} SPARSE~\cite{lustig2006kt}, while being computationally more appealing as it relies on a successive quadratic optimization. For TSL, we choose $R=75$ and allow $4$ epochs, while for the {\it k-t} FOUCSS we use the publicly available source code provided in~\cite{jung2009k} with $2$ outer iterations and $40$ inner iterations. Runtime (seconds) is compared for the {\it k-t} FOCUSS and TSL schemes in Table~\ref{tab:table_runtime_batch}, where it is evident that the TSL scheme converges much faster especially when running on GPU. Note that {\it k-t} FOCUSS cannot be implemented on GPU for the provided code since each iteration involves the large complex-valued tensor $\ubX \in \mathbbm{C}^{200 \times 256 \times 256}$, which exceeds the GPU memory.

\begin{table}[t]
	\caption{Runtime (sec) and average NMSE for batch processing of {\it k-t} TSL and FOCUSS~\cite{jung2009k} under $10$-fold acceleration.}
	\vspace{-2.5mm}
	\label{tab:table_runtime_batch}
	\begin{center}
		\begin{tabular} {|c|c|c|c|}
			\hline
			Scheme &  Avg. NMSE & Runtime (CPU) & Runtime (GPU)  \\
			\hline\hline
			FOCUSS & $0.049$ & $340$ & N/A \\
			\hline
			TSL & $0.01$ & $84$ & $27.75$ \\
			\hline
		\end{tabular}
	\end{center}
\end{table}

\subsection{Tomographic parallel MRI}
\label{subsec:parallel_mri_test}
\begin{figure}[t]
	\centering
	\includegraphics[scale=1.5]{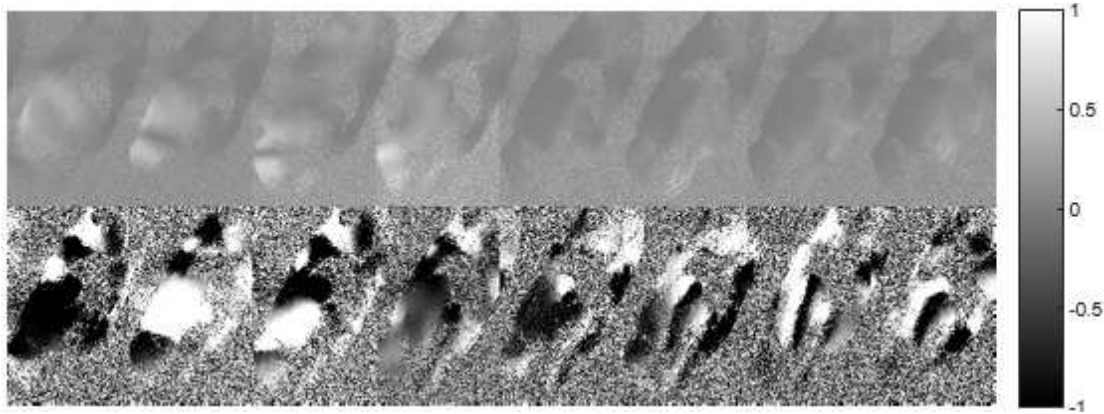} 
	\caption{Coil sensitivity magnitude (top) and phase (bottom) for eight coils in parallel MRI dataset D2 with $R=200$ and $\lambda=1$. The phase is normalized by $\pi$.}
	\label{fig:fig_coil_sens}
\end{figure}

\begin{figure}[t]
	\centering
	\includegraphics[scale=1.20]{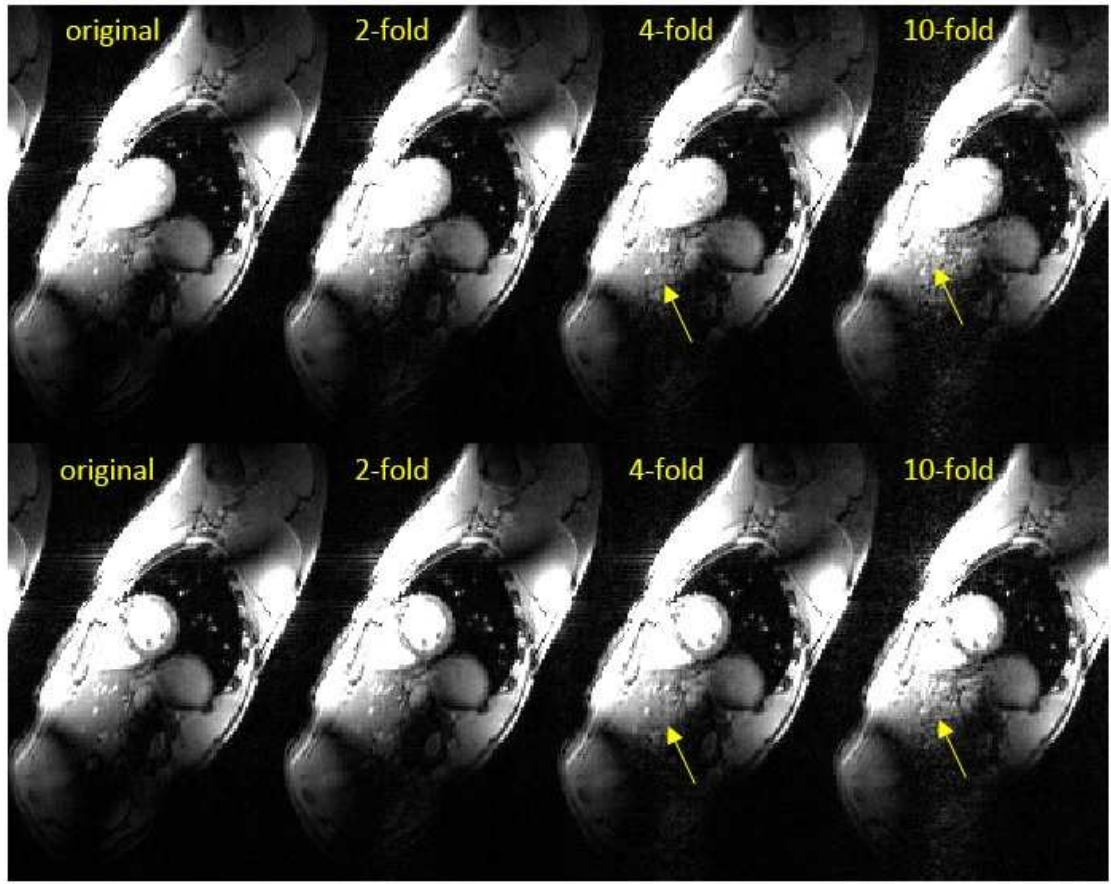}
	\caption{Reconstructed parallel MRI images under different acceleration factors for two representative images corresponding to top and bottom rows when $R=200$ and $\lambda=1$. }
	\label{fig:fig_recon_image_parallel}
\end{figure}

The multi-coil dataset (D2) is used to evaluate performance for parallel MRI reconstruction. The coil sensitivity maps are estimated using the sum-of-squares method from the first $k$-space frame that is acquired fully. Phase smoothing via polynomial fitting as in \cite{pruessmann1999sense} is then used to smooth the sharp phase transitions. The resulting phase and magnitude for the first $8$ coils are depicted in Fig.~\ref{fig:fig_coil_sens}. Tensor rank was chosen to be $R=200$. Due to limited number of temporal slices ($T=26$), the batch procedure is adopted with multiple visits over data. Selecting $\lambda=1$, two representative reconstructed images under various acceleration factors are depicted in Fig.~\ref{fig:fig_recon_image_parallel}. Apparently, the reconstructed images qualitatively look quite close to the ground-truth one, where for rate $10$ acceleration some flow artifacts (as pinpointed by arrows) begin appearing around the heart. This is also quantitatively confirmed by the small average NMSE listed in Table~\ref{tab:table_nmse_parallel}.

\begin{table}[t]
	\caption{NMSE of the tomographic parallel MRI under various accleration factors when $R=200$.}
	\vspace{-2.5mm}
	\label{tab:table_nmse_parallel}
	\begin{center}
		\begin{tabular} {|c|c|c|c|c|}
			\hline
			Acceleration &  1 & 2 & 4 & 10  \\
			\hline\hline
			NMSE & $0.0029$ & $0.0031$ & $0.0038$ & $0.0067$ \\
			\hline
		\end{tabular}
	\end{center}
\end{table}

\section{Conclusions and future directions} 
\label{sec:conc}
Fast and effective analytics are developed in this paper for real-time reconstruction of MR images. Treating a temporal image sequence as a multi-way data array, a novel tensor subspace learning approach was introduced based on the PARAFAC decomposition. The correlation structure across various dimensions is learned from the tensor's latent subspace. A rank-regularized least-squares estimator was put forth to learn this tensor subspace when a Tykhonov-type regularizer which promotes low rank for the data tensor. Adopting stochastic alternating minimization, recursive algorithms were developed to track the subspace and reconstruct the image 'on the fly.' The resulting algorithm enjoys lightweight iterations with parallelized computations, which makes it attractive for high-resolution real-time MRI. Leveraging the online subspace iterates offered by the algorithm, adaptive subsampling strategies were also devised to randomly predict informative samples for acquisition of subsequent $k$-space frames. GPU-based simulated tests with real cardiac cine MRI datasets corroborated the effectiveness of the novel approach. To broaden the scope of the present work there are several intriguing questions to address including the extension and evaluation of the novel analytics for volumetric and 4D MRI with more sophisticated such as radial sampling trajectories. Another important direction pertains to performance analysis of the adaptive subsampling policy for selecting the best set of $k$-space samples.

\section{Acknowledgments}
We would like to acknowledge Dr. Sebatian Schmitter from the Center for Magnetic Resonance Research, Minneapolis, Minnesota, for providing us with the dataset (D2) and helpful discussions. The valuable comments by Dr. Leslie Ying from the State University of New York at Buffalo  who provided us with the dataset (D2) are also gratefully appreciated.

\bibliographystyle{IEEEtranS}
\bibliography{IEEEabrv,biblio}

\end{document}